\documentclass{article}

\usepackage[final,nonatbib]{neurips_2023}

\usepackage{graphicx}
\usepackage{hyperref}  
\usepackage[utf8]{inputenc} 
\usepackage[T1]{fontenc}    
\usepackage{url}            
\usepackage{booktabs}       
\usepackage{amsfonts}       
\usepackage{nicefrac}       
\usepackage{microtype}      
\usepackage{xcolor}         
\usepackage{algorithm}
\usepackage{algorithmic}
\usepackage{tikz}
\usepackage{wrapfig}
\usepackage{adjustbox}

\usepackage{amsmath}
\usepackage{amssymb}
\usepackage{mathtools}
\usepackage{amsthm}

\usetikzlibrary{bayesnet}  
\usetikzlibrary{positioning}  

\usepackage[nameinlink,capitalize,noabbrev]{cleveref}

\theoremstyle{plain}
\newtheorem{theorem}{Theorem}[section]
\newtheorem{proposition}[theorem]{Proposition}
\newtheorem{lemma}[theorem]{Lemma}

\theoremstyle{definition}

\theoremstyle{remark}

\usepackage[textsize=tiny]{todonotes}

\usepackage{amsthm, amssymb}
\usepackage{hyperref}
\usepackage{booktabs}
\usepackage{comment}
\usepackage{graphicx}
\newif\ifshowcomments
\showcommentsfalse
\usepackage{marginnote}
\usepackage{bbm}
\usepackage{wrapfig}
\usepackage{caption}
\usepackage{subcaption}
\usepackage{adjustbox}

\newcommand{\ModelName}{\texttt{BayesDAG}} 

\usepackage{wrapfig}
\usepackage{alphalph}
\usepackage[shortlabels]{enumitem}

\usepackage{amsmath,amsfonts,bm}

\def\eqref#1{equation~\ref{#1}}

\def\1{\bm{1}}

\def\vf{{\bm{f}}}

\def\vo{{\bm{o}}}
\def\vp{{\bm{p}}}

\def\vr{{\bm{r}}}

\def\vu{{\bm{u}}}

\def\vx{{\bm{x}}}

\def\mD{{\bm{D}}}
\def\mE{{\bm{E}}}

\def\mG{{\bm{G}}}

\def\mL{{\bm{L}}}
\def\mM{{\bm{M}}}

\def\mV{{\bm{V}}}
\def\mW{{\bm{W}}}
\def\mX{{\bm{X}}}

\def\tmW{{\tilde{\bm{W}}}}

\DeclareMathAlphabet{\mathsfit}{\encodingdefault}{\sfdefault}{m}{sl}
\SetMathAlphabet{\mathsfit}{bold}{\encodingdefault}{\sfdefault}{bx}{n}

\def\sD{{\mathbb{D}}}

\def\sR{{\mathbb{R}}}

\def\Pa{{\text{\textbf{Pa}}}}

\newcommand{\E}{\mathbb{E}}

\newcommand{\KL}{D_{\mathrm{KL}}}

\DeclareMathOperator*{\argmax}{arg\,max}

\DeclareMathOperator*{\grad}{grad}
\DeclareMathOperator*{\relu}{ReLU}
\DeclareMathOperator*{\step}{Step}
\DeclareMathOperator*{\indicator}{\mathbbm{1}}

\DeclareMathOperator{\sampler}{Sampler}

\DeclareMathOperator{\ELBO}{ELBO}

\newcommand{\binaryset}{\{0,1\}^{d\times d}}
\def\mTheta{{\bm{\Theta}}}

\makeatletter
\newtheorem*{rep@theorem}{\rep@title}
\newcommand{\newreptheorem}[2]{%
\newenvironment{rep#1}[1]{%
 \def\rep@title{#2 \ref{##1}}%
 \begin{rep@theorem}}%
 {\end{rep@theorem}}}
\makeatother

\newreptheorem{theorem}{Theorem}
\def\perm{{\bm{\sigma}}}
\def\Sinkhorn{{\mathcal{S}}}
\def\polytope{{\mathcal{B}_d}}

\title{BayesDAG:  Gradient-Based Posterior Inference for Causal Discovery}

\makeatletter
\newcommand{\printfnsymbol}[1]{%
  \textsuperscript{\@fnsymbol{#1}}%
}
\makeatother
\author{%
\textbf{Yashas Annadani}$^{\printfnsymbol{2}\thanks{Equal contribution. \printfnsymbol{2} Work done during internship at Microsoft Research. Correspondence to \texttt{wenbogong@microsoft.com}. Code: \url{https://github.com/microsoft/Project-BayesDAG}}\,\,\,1,3,4}$
\quad
\textbf{Nick Pawlowski}$^{2}$
\quad
\textbf{Joel Jennings}$^{2}$
\quad
\textbf{Stefan Bauer}$^{3, 4}$
\\
\textbf{Cheng Zhang}$^{2}$
\quad
\textbf{Wenbo Gong}$^{\printfnsymbol{1}2}$
\\
$^1$ KTH Royal Institute of Technology, Stockholm \quad $^2$ Microsoft Research\\ $^3$ Helmholtz AI, Munich $^4$ TU Munich\\
}

\begin{document}

\maketitle

\begin{abstract}
\looseness=-1 Bayesian causal discovery aims to infer the posterior distribution over causal models from observed data, quantifying epistemic uncertainty and benefiting downstream tasks. However, computational challenges arise due to joint inference over combinatorial space of Directed Acyclic Graphs (DAGs) and nonlinear functions. Despite recent progress towards efficient posterior inference over DAGs,  existing methods are either limited to variational inference on node permutation matrices for linear causal models, leading to compromised inference accuracy, or continuous relaxation of adjacency matrices constrained by a DAG regularizer, which cannot ensure resulting graphs are DAGs. In this work, we introduce a scalable Bayesian causal discovery framework based on a combination of stochastic gradient Markov Chain Monte Carlo (SG-MCMC) and Variational Inference (VI) that overcomes these limitations. Our approach directly samples DAGs from the posterior without requiring any DAG regularization, simultaneously draws function parameter samples and is applicable to both linear and nonlinear causal models. To enable our approach, we derive a novel equivalence to the permutation-based DAG learning, which opens up possibilities of using any relaxed gradient estimator defined over permutations. To our knowledge, this is the first framework applying gradient-based MCMC sampling for causal discovery. Empirical evaluation on synthetic and real-world datasets demonstrate our approach's effectiveness compared to state-of-the-art baselines.
\end{abstract}

\section{Introduction}
\label{sec:introduction}
The quest for discovering causal relationships in data-generating processes lies at the heart of empirical sciences and decision-making~\cite{pe2001inferring,sachs2005causal, van2006application}. Structural Causal Models (SCMs) \cite{pearl2009causality} and their associated Directed Acyclic Graphs (DAGs) provide a robust mathematical framework for modeling such relationships. Knowledge of the underlying SCM and its corresponding DAG permits predictions of unseen interventions and causal reasoning, thus making causal discovery -- learning an unknown SCM and its associated DAG from observed data -- a subject of extensive research \cite{peters2017elements,spirtes2000causation}.

In contrast to traditional methods that infer a single graph or its Markov equivalence class (MEC) \cite{chickering2002optimal, spirtes2000causation},  Bayesian causal discovery~\cite{friedman2003being,heckerman2006bayesian,tong2001active} aims to infer a posterior distribution over SCMs and their DAGs from observed data. This approach encapsulates the epistemic uncertainty, degree of confidence in every causal hypothesis, which is particularly valuable for real-world applications when data is scarce. It is also beneficial for downstream tasks such as experimental design \cite{agrawal2019abcd,annadani2023differentiable,murphy2001active, tigas2022interventions}.

\looseness=-1 The central challenge in Bayesian causal discovery lies in inferring the posterior distribution over the union of the exponentially growing (discrete) DAGs and (continuous) 
function parameters. Prior works have used Markov Chain Monte Carlo (MCMC) to directly sample DAGs or bootstrap traditional discovery methods \cite{chickering2002optimal, murphy2001active, tong2001active}, but these methods are typically limited to linear models which admit closed-form marginalization over continuous parameters. Recent advances have begun to utilize gradient information for more efficient inference. These approaches are either: (1) DAG regularizer-based methods, e.g.~ DIBS \cite{lorch2021dibs}, which use continuous relaxation of adjacency matrices together with DAG regularizer \cite{zheng2018dags}. But DIBS formulation fails to model edge co-dependencies and suffer from inference quality due to its inference engine (Stein variational gradient descent) \cite{gong2020sliced, gong2021active}. Additionally, all DAG regularizer based methods cannot guarantee DAG generation; (2) permutation-based DAG learning, which directly infers permutation matrices and guarantees to generate DAGs. However, existing works focus on using only variational inference \cite{charpentier2022differentiable,cundy2021bcd}, which may suffer from inaccurate inference quality \cite{gong2019icebreaker,springenberg2016bayesian, trippe2018overpruning} and is sometimes restricted to only linear models \cite{cundy2021bcd}. 

In this work, we introduce \ModelName, a gradient-based Bayesian causal discovery framework that overcomes the above limitations. Our contributions are:
\begin{enumerate}
\item We prove that an augmented space of edge beliefs and node potentials $(\mW,\vp)$, similar to NoCurl \cite{yu2021dags}, permits equivalent Bayesian inference in DAG space without the need for any regularizer. (\cref{subsec: Bayesian inference W p space})

\item We derive an equivalence relation from this augmented space to permutation-based DAG learning which provides a general framework for gradient-based posterior inference. (\cref{subsec: equivalent formulation})

\item Based on this general framework, we propose a scalable Bayesian causal discovery that is model-agnostic for linear and non-linear cases and also offers improved inference quality. We instantiate our approach through two formulations: (1) a combination of SG-MCMC and VI  (2) SG-MCMC with a continuous relaxation. (\cref{sec: SGMCMC sampling framework})

\item We demonstrate the effectiveness of our approach in providing accurate Bayesian inference quality and superior causal discovery performance with comprehensive empirical evaluations on various datasets. We also demonstrate that our method can be easily scaled to $100$ variables with nonlinear relationships. (\cref{sec: experiments})
\end{enumerate}

\section{Background}
\label{sec: Prerequisite}

\paragraph{Causal Graph and Structural Causal Model}
Consider a data generation process with $d$ variables $\mX\in\sR^{d}$. The causal relationships among these variables is  represented by a Structural Causal Model (SCM) which consists of a set of structural equations~\cite{peters2017elements} where each variable $X_i$ is a function of its direct causes $\mX_{\Pa^i}$ and an exogenous noise variable $\epsilon_i$ with distribution $P_{\epsilon_i}$: 
\begin{equation}
X_i\coloneqq f_i(\mX_{\Pa^i},\epsilon_i)
\label{eq: CM}
\end{equation}
These equations induce a 
causal graph $\mG=(\mV,\mE)$, comprising a node set $\mV$ with $\vert\mV\vert=d$ indexing the variables $\mX$ and a directed edge set $\mE$. If a directed edge $e_{ij}\in\mE$ exists between a node pair $v_i,v_j\in\mV$ (i.e., $v_i\rightarrow v_j$), we say that $X_i$ causes $X_j$ or $X_i$ is the parent of $X_j$. We use the binary adjacency matrix $\mG\in\binaryset$ to represent the causal graph, where the entry $G_{ij}=1$ denotes $v_i \rightarrow v_j$. 
A standard assumption in causality is that the structural assignments are acyclic and the induced causal graph is a DAG~\cite{bongers2021foundations, pearl2009causality}, which we adopt in this work.
We further assume that the SCM is causally sufficient i.e. all variables are measurable and exogenous noise variables $\epsilon_i$ are mutually independent. Throughout this work, we consider a special form of SCM called Gaussian additive noise model (ANM):
\begin{equation}
X_i\coloneqq f_i(\mX_{\Pa^i}) + \epsilon_i~~~~~~\text{where}~~~\epsilon_i\sim\mathcal{N}(0,\sigma^2_i)
\label{eq: ANM}
\end{equation}
If the functions are not linear or constant in any of its arguments, the Gaussian ANM is structurally identifiable~\cite{hoyer2008nonlinear,peters2014causal}.

\paragraph{Bayesian Causal Discovery}
Given a dataset $\mD=\{\vx^{(1)},\ldots,\vx^{(N)}\}$ with i.i.d observations, underlying graph $\mG$ and SCM parameters $\mTheta$
, they induce a unique joint distribution $p(\mD,\mTheta, \mG) = p(\mD\vert \mG,\mTheta)p(\mG,\mTheta)$ with the prior $p(\mG,\mTheta)$ and likelihood $p(\mD\vert \mG,\mTheta)$ \cite{friedman2003being}. Under finite data and/or limited identifiability of SCM (e.g upto MEC), it is desirable to have accurate uncertainty estimation for downstream decision making rather than inferring a single SCM and its graph (for e.g. with a maximum likelihood estimate). Bayesian causal discovery therefore aims to infer the posterior $p(\mG,\mTheta\vert \mD)=p(\mD,\mTheta, \mG)/p(\mD)$. However, this posterior is intractable due to the super-exponential growth of the possible DAGs $\mG$~\cite{robinson1973counting} and continuously valued model parameters $\mTheta$ in nonlinear functions. VI \cite{zhang2018advances} or SG-MCMC \cite{gong2022advances,ma2015complete} are two types of methods developed to tackle general Bayesian inference problems, but adaptations are required for Bayesian causal discovery. 

\paragraph{NoCurl Characterization}
Inferring causal graphs is challenging due to the DAG constraint. Previous works \cite{geffner2022deep, gong2022rhino,lachapelle2019gradient,lorch2021dibs,yu2019dag} directly infer adjacency matrix with the DAG regularizer~\cite{zheng2018dags}. However, it requires an annealing schedule, resulting in slow convergence, and no guarantees on generating DAGs. Recently, \cite{yu2021dags} introduced NoCurl, a novel characterization of the \textbf{weighted DAG} space. They define a potential $p_i\in \sR$ for each node $i$, grouped as potential vector $\vp\in\sR^d$ . Further, a gradient operator on $\vp$ mapping it to a skew-symmetric matrix is introduced:
\begin{equation}
    (\grad \vp)(i,j) = p_i-p_j
    \label{eq: grad operator}
\end{equation}
Based on the above operation, a mapping that directly maps from the augmented space $(\mW,\vp)$ to the DAG space $\gamma(\cdot,\cdot):\sR^{d\times d}\times \sR^{d}\rightarrow \sR^{d\times d}$ was proposed:
\begin{equation}
\gamma(\mW,\vp) = \mW\odot \relu(\grad \vp)
\label{eq: NoCurl mapping}
\end{equation}
where $\relu(\cdot)$ is the ReLU activation function and $\mW$ is a skew-symmetric \textbf{continuously weighted} matrix. This formulation is complete (Theorem 2.1 in \cite{yu2021dags}), as any continuously weighted DAG can be represented by a $(\mW,\vp)$ pair and vice versa.
NoCurl translates the learning of a single weighted DAG to a corresponding $(\mW,\vp)$ pair. However, direct gradient-based optimization is challenging due to a highly non-convex loss landscape, which leads to the reported failure in \cite{yu2021dags}. 

Although NoCurl appears suitable for our purpose, the failure in directly learning suggests non-trivial optimizations. We hypothesize that this arises from the continuously weighted matrix $\mW$. In the following, we introduce our proposed parametrization inspired by NoCurl to characterize the \textbf{binary} DAG adjacency matrix.

\section{Sampling the DAGs}
\label{sec: sampling the DAGs}
In this section, we focus on the Bayesian inference over binary DAGs through a novel mapping, $\tau(\mW,\vp)$, a modification of NoCurl. We establish the validity of performing Bayesian inference within $(\mW,\vp)$ space utilizing $\tau$ (\cref{subsec: Bayesian inference W p space}). However, $\tau$ yields uninformative gradient during back-propagation, a challenge we overcome by deriving an equivalent formulation based on permutation-based DAG learning, thereby enabling the use of relaxed gradient estimators (\cref{subsec: equivalent formulation}).

\subsection{Bayesian Inference in $W,p$ Space}
\label{subsec: Bayesian inference W p space}
The NoCurl formulation (\cref{eq: NoCurl mapping}) focuses on learning \emph{a single weighted} DAG, which is not directly useful for our purpose. We need to address two key questions: (1) considering only binary adjacency matrices without weights; (2) ensuring Bayesian inference in $(\mW,\vp)$ is valid.

We note that the proposed transformation in NoCurl $\gamma$ (\cref{eq: NoCurl mapping} can be hard to optimize for the following reasons: (i) $\relu(\grad \vp)$ 
 gives a fully connected DAG. The main purpose of $\mW$ matrix therefore is to disable the edges. Continuous $\mW$ requires thresholding to properly disable the edges, since it is hard for a continuous matrix to learn exactly $0$ during the optimization; (ii) 
 $\relu(\grad \vp)$ and $\mW$
 are both continuous valued matrices. Thus, learning of the edge weights and DAG structure are not explicitly separated, resulting in complicated non-convex optimizations\footnote{See discussion below Eq. 3 in~\cite{yu2021dags} for more details.}. Parameterizing the search space in terms of binary adjacency matrices significantly simplifies the optimization complexity as the aforementioned issues are circumvented. Therefore, we introduce a modification $\tau:\binaryset\times\sR^d\rightarrow \binaryset$:
\begin{equation}
\tau(\mW,\vp) = \mW\odot \step(\grad \vp)
\label{eq: Binary NoCurl}
\end{equation}
where we abuse the term $\mW$ for binary matrices, and replace $\relu(\cdot)$ with $\step(\cdot)$. $\mW$ acts as mask to disable the edge existence. Thus, due to the $\step$, $\tau$ can only output a binary adjacency matrix.

Next, we show that performing Bayesian inference in such augmented $(\mW,\vp)$ space is valid, i.e., using the posterior $p(\mW,\vp\vert \mD)$ to replace $p(\mG\vert \mD)$. This differs from NoCurl, which focuses on a single graph rather than the validity for Bayesian inference, requiring a new theory for soundness.

\begin{theorem}[Equivalence of inference in $(\mW,\vp)$ and binary DAG space]
Assume graph $\mG$ is a binary adjacency matrix representing a DAG and node potential $\vp$ does not contain the same values, i.e.~$p_i\neq p_j$ $\forall i,j$. Then, with the induced joint observational distribution $p(\mD,\mG)$, dataset $\mD$, and a corresponding prior $p(\mG)$, we have
\begin{align}
p(\mG\vert \mD) = \int p_\tau(\vp,\mW\vert \mD)\indicator(\mG=\tau(\mW,\vp))d\mW d\vp
\label{eq: equivalence of bayesian inference}
\end{align}
if $p(\mG)=\int p_\tau(\vp,\mW)\indicator(\mG=\tau(\mW,\vp))d\mW d\vp$, where $p_\tau(\mW,\vp)$ is the prior, $\indicator(\cdot)$ is the indicator function, and $p_\tau(\vp,\mW\vert D)$ is the posterior distribution over $\vp,\mW$.
\label{thm: equivalence of bayesian inference}
\end{theorem}

Refer to \cref{subapp: proof of equivalence of bayesian inference} for detailed proof.

This theorem guarantees that instead of performing inference directly in the constrained space (i.e.~DAG space), we can apply Bayesian inference in a less complex $(\mW,\vp)$ space where $\mW\in\binaryset$ and $\vp\in\sR^d$ without explicit constraints.

For inference of $\vp$, we adopt a sampling-based approach, which is asymptotically accurate \cite{ma2015complete}. In particular, we consider SG-MCMC (refer to \cref{sec: SGMCMC sampling framework}), which avoids the expensive Metropolis-Hastings acceptance step and scales to large datasets. We emphasize that any other suitable sampling algorithms can be directly plugged in, thanks to the generality of the framework.

However, the mapping $\tau$ does not provide meaningful gradient information for $\vp$ due to the piecewise constant $\step(\cdot)$ function, which is required by SG-MCMC.


\subsection{Equivalent Formulation}
\label{subsec: equivalent formulation}
\looseness=-1 In this section, we address the above issue by deriving an equivalence to a permutation learning problem. This alternative formulation enables various techniques that can approximate the gradient of $\vp$. 

\paragraph{Intuition} The node potential $\vp$ implicitly defines a topological ordering through the mapping $\step(\grad(\cdot))$. In particular, $\grad(\cdot)$ outputs a skew-symmetric adjacency matrix, where each entry specifies the potential difference between nodes. $\step(\grad(\cdot))$ zeros out the negative potential differences (i.e.~$p_i\leq p_j$), and only permits the edge direction from higher potential to the lower one (i.e.~$p_i>p_j$). This implicitly defines a sorting operation based on the descending node potentials, which can be cast as a particular $\argmax$ problem \cite{blondel2020fast,kuhn1955hungarian, mena2018learning, niculae2018sparsemap,zantedeschi2023dag} involving a permutation matrix.

\paragraph{Alternative formulation} We define $\mL\in\binaryset$ as a matrix with lower triangular part to be $1$, and vector $\vo=[1,\ldots, d]$. We propose the following formulation:
\begin{align}
&\mG = \mW \odot \left[\perm(\vp)\mL \perm(\vp)^T\right] \
&\text{where}\; \perm(\vp) = \argmax_{\perm'\in\bm{\Sigma}_d} \vp^T(\perm' \vo)
\label{eq: alternative formulation}
\end{align}
Here, $\bm{\Sigma}_d$ represents the space of all $d$ dimensional permutation matrices. The following theorem states the equivalence of this formulation to  \cref{eq: Binary NoCurl}.
\begin{theorem}[Equivalence to NoCurl formulation]
Assuming the conditions in \cref{thm: equivalence of bayesian inference} are satisfied. Then, for a given $(\mW,\vp)$, we have
$$
\mG=\mW\odot \step(\grad \vp) = \mW\odot \left[ \perm(\vp)\mL\perm(\vp)^T\right]
$$
where $\mG$ is a DAG and $\perm(\vp)$ is defined in \cref{eq: alternative formulation}.
\label{thm: equivalence of alternative formulation}
\end{theorem}
Refer to \cref{subapp: proof of theorem alternative formulation} for details.

This theorem translates our proposed operator $\step(\grad(\vp))$ into finding a corresponding permutation matrix $\perm(\vp)$. Although this does not directly solve the uninformative gradient, it opens the door for approximating this gradient with the tools from the differentiable permutation literature \cite{blondel2020fast,mena2018learning, niculae2018sparsemap}. For simplicity, we adopt the Sinkhorn approach \cite{mena2018learning}, but we emphasize that this equivalence is general enough that any past or future approximation methods can be easily applied.

\paragraph{Sinkhorn operator} The Sinkhorn operator $\Sinkhorn(\mM)$ on a matrix $\mM$ \cite{adams2011ranking} is defined as a sequence of row and column normalizations, each is called Sinkhorn iteration.

\cite{mena2018learning} showed that the non-differentiable $\argmax$ problem
\begin{equation}
    \perm = \argmax_{\perm'\in\Sigma_d}\left\langle\perm', \mM\right\rangle
    \label{eq: permutation argmax}
\end{equation}
\looseness=-1 can be relaxed through an entropy regularizer with its solution being expressed by $\Sinkhorn(\cdot)$.
In particular, they showed that $\Sinkhorn(\mM/t)=\argmax_{\perm'\in \polytope}\left\langle\perm',\mM\right\rangle+th(\perm')$, where $h(\cdot)$ is the entropy function. This regularized solution converges to the solution of \cref{eq: permutation argmax} when $t\rightarrow 0$, i.e.~$\lim_{t\rightarrow 0}\Sinkhorn(\mM/t)$. 
Since the Sinkhorn operator is differentiable, $\Sinkhorn(\mM/t)$ can be viewed as a differentiable approximation to \cref{eq: permutation argmax}, which can be used to obtain the solution of \cref{eq: alternative formulation}. Specifically, we have
\begin{equation}
    \argmax_{\perm'\in\bm{\Sigma}_d} \vp^T(\perm'\vo) = \argmax_{\perm'\in\bm{\Sigma}_d}\langle\perm',\vp\vo^T\rangle = \lim_{t\rightarrow 0}\Sinkhorn(\frac{\vp\vo^T}{t})
    \label{eq: Sinkhorn solution}
\end{equation}
In practice, we approximate it wth $t>0$, resulting in a doubly stochastic matrix. To get the binary permutation matrix, we apply the Hungarian algorithm \cite{munkres1957algorithms}. During the backward pass, we use a straight-through estimator~\cite{bengio2013estimating} for $\vp$. 

Some of the previous works \cite{charpentier2022differentiable,cundy2021bcd} have leveraged the Sinkhorn operator to model variational distributions over permutation matrices. However, they start with a full rank $\mM$, which has been reported to require over \textbf{1000} Sinkho    rn iterations to converge \cite{cundy2021bcd}. However, our formulation, based on explicit node potential $\vp\vo^T$, generates a rank-1 matrix, requiring much fewer Sinkhorn steps (around \textbf{300}) in practice, saving two-thirds of the computational cost.

\section{Bayesian Causal Discovery via Sampling}
\label{sec: SGMCMC sampling framework}
In this section, we delve into two specific methodologies that are derived from the proposed framework. The first one, which will be our main focus, combines SG-MCMC and VI in a Gibbs sampling manner. The second one, which is based entirely on SG-MCMC with continuous relaxation, is also derived, but we include its details in \cref{appsec: joint inference SG-MCMC} due to its inferior empirical performance.

\subsection{Model Formulation}
\label{subsect: model formulation}
We build upon the model formulation of \cite{geffner2022deep}, which combines the additive noise model with neural networks to describe the functional relationship. Specifically, $X_i \coloneqq f_i(\mX_{\Pa^i}) + \epsilon_i$, where $f_i$ adheres to the adjacency relation specified by $\mG$, i.e.~$\partial f_i(\vx) / \partial x_j = 0$ if no edge exists between nodes $i$ and $j$. We define $f_i$ as 
\begin{equation}
    f_i(\vx) = \zeta_i\left(\sum_{j=1}^dG_{ji}l_j(x_j)\right),
\end{equation}
where $\zeta_i$ and $l_i$ are neural networks with parameters $\mTheta$, and $\mG$ serves as a mask disabling non-parent values. To reduce the number of neural networks, we adopt a weight-sharing mechanism: $\zeta_i(\cdot) = \zeta(\vu_i,\cdot)$ and $l_i(\cdot) = l(\vu_i,\cdot)$, with trainable node embeddings $\vu_i$.

\paragraph{Likelihood of SCM}
The likelihood can be evaluated through the noise $\bm{\epsilon} = \vx - \vf(\vx;\mTheta)$. \cite{geffner2022deep} showed that if $\mG$ is a DAG, then the mapping from $\bm{\epsilon}$ to $\vx$ is invertible with a Jacobian determinant of 1. Thus, the observational data likelihood is:
\begin{equation}
    p(\vx \vert \mG) = p_\epsilon(\vx-\vf(\vx;\mTheta)) = \prod_{i=1}^d p_{\epsilon_i}(x_i-f_i(\vx_{\Pa_G^i}))
    \label{eq: SEM likelihood}
\end{equation}
\paragraph{Prior design}
We implicitly define the prior $p(\mG)$ via $p(\vp,\mW)$. We propose the following for the joint prior:
\begin{equation}
    p(\mW,\vp, \mTheta) \propto \nonumber \mathcal{N}(\mTheta;\bm{0},\bm{I})  \mathcal{N}(\vp;\bm{0},\alpha\bm{I})\mathcal{N}(\mW;\bm{0},\bm{I})\exp(-\lambda_s\Vert\tau(\mW,\vp)\Vert^2_F)
    \label{eq: Prior p w}
\end{equation}
where $\alpha$ controls the initialization scale of $\vp$ and $\lambda_s$ controls the sparseness of $\mG$. 

\subsection{Bayesian Inference of $W,p,\Theta$}
\label{subsubsec: combined inference}

The main challenge lies in the binary nature of $\mW \in \{0,1\}^{d\times d}$, which requires a discrete sampler. Although recent progress has been made \cite{grathwohl2021oops,sun2022discrete,zanella2020informed,zhang2022langevin}, these methods either involve expensive Metropolis-Hasting (MH) steps or require strong assumptions on the target posterior when handling batched gradients. To address this, we propose a combination of SG-MCMC for $\vp,\mTheta$ and VI for $\mW$. It should be noted that our framework can incorporate any suitable discrete sampler if needed. 
\begin{wrapfigure}[13]{r}{0.25\textwidth}
    \vspace{2em}
    \usetikzlibrary{bayesnet}
\scalebox{0.6}{
\begin{tikzpicture}[scale=3]

  \node[latent] (W) {$\mW$};
  \node[latent, left=of W] (p) {$\vp$};
  \node[latent, below=of W, xshift=1.2cm] (Theta) {$\mTheta$};
  \node[obs, below=of Theta] (x) {$\vx^{(i)}$};
  
  \edge {p} {x};
  \edge {W} {Theta};
  \edge {W} {x};
  \edge {Theta} {x};
  \edge {p} {Theta};
  
  \plate {x_plate} {(x)} {$i = 1, \dots, N$};

\end{tikzpicture}
}
    \caption{Graphical model of the inference problem.}
    \label{fig: graphical model with latent}
\end{wrapfigure}
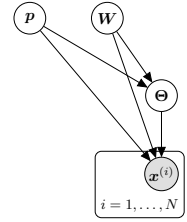

We employ a Gibbs sampling procedure \cite{casella1992explaining}, which iteratively applies (1) sampling $\vp,\mTheta\sim p(\vp,\mTheta|\mD,\mW)$ with SG-MCMC; (2) updating the variational posterior $q_\phi(\mW|\vp,\mD)\approx p(\mW|\vp,\mTheta,\mD)$.

We define the posterior $p(\vp,\mTheta \vert \mD,\mW)\propto \exp(-U(\vp,\mW,\mTheta))$, where $U(\vp,\mW,\mTheta) = -\log p(\vp,\mD,\mW,\mTheta)$. SG-MCMC in continuous time defines a specific form of It\^o diffusion that maintains the target distribution invariant \cite{ma2015complete} without the expensive computation of the MH step. We adopt the Euler-Maruyama discretization for simplicity. Other advanced discretization can be easily incorporated \cite{chen2015convergence,platen2010numerical}. 

Preconditioning techniques have been shown to accelerate SG-MCMC convergence \cite{chen2014stochastic, gong2019icebreaker,li2016preconditioned,welling2011bayesian,ye2017langevin}. We modify the sampler based on \cite{gong2019icebreaker}, which is inspired by Adam \cite{kingma2014adam}. Detailed update equations can be found in \cref{app: SG-MCMC update}.

The following proposition specifies the gradients required by SG-MCMC: $\nabla_{\vp,\mTheta} U(\vp,\mW,\mTheta)$.


\begin{proposition}
Assume the model is defined as above, then we have the following:
\begin{equation}
    \nabla_{\vp} U=-\nabla_{\vp}\log p(\vp) - \nabla_{\vp}\log p(\mD\vert \mTheta, \tau(\mW,\vp))
    \label{eq: p gradient}
\end{equation}
and 
\begin{equation}
    \nabla_{\mTheta} U =-\nabla_{\mTheta}\log p(\mTheta) - \nabla_{\mTheta}\log p(\mD\vert \mTheta,\tau(\vp,\mW))
    \label{eq: mTheta gradient}
\end{equation}
\label{prop: gradient computation}
\end{proposition}
Refer to \cref{subapp: proof of proposition gradient computation} for details.

\paragraph{Variational inference for $\mW$}
We use the variational posterior $q_\phi(\mW\vert \vp)$ to approximate the true posterior $p(\mW\vert \vp,\mTheta,\mD)$. Specifically, we select an independent Bernoulli distribution with logits defined by the output of a neural network $\mu_\phi(\vp)$:
\begin{equation}
    q_\phi(\mW\vert \vp)=\prod_{ij}Ber(\mu_\phi(\vp)_{ij})
    \label{eq:VI bernoulli}
\end{equation}
To train $q_\phi$, we derive the corresponding \emph{evidence lower bound} (ELBO):
\begin{equation}
    \ELBO(\phi) = \E_{q_\phi(\mW|\vp)}\left[
    \log p(\mD,\vp,\mTheta\vert \mW)]-\KL\left[q_\phi(\mW\vert \vp)\Vert p(\mW)\right]
    \right].
    \label{eq: ELBO for W}
\end{equation}
where $\KL$ is the Kullback-Leibler divergence. 
The derivation is in \cref{subapp: derivation of ELBO}.
\cref{alg: combined inference} summarizes this inference procedure. 
\begin{algorithm}[tb]
\caption{\ModelName~ SG-MCMC+VI Inference}
\label{alg: combined inference}
\begin{algorithmic}
\STATE {\bfseries Input:} dataset $\mD$; prior $p(\vp, \mW),p(\mTheta)$; SG-MCMC sampler $\sampler$; sampler hyperparameters $\Psi$; network $\mu_\phi(\cdot)$; training iteration $T$.
\STATE {\bfseries Output:} samples $\{\mTheta,\vp\}$ and variational posterior $q_\phi$
\STATE Initialize $\mTheta^{(0)}, \vp^{(0)},\phi$
\FOR{$t=1\ldots,T$}
    \STATE Sample $\mW^{(t-1)}\sim q_{\phi}(\mW\vert \vp^{(t-1)})$
    \STATE Evaluate $\nabla_{\vp,\mTheta} U$ (\cref{eq: p gradient,eq: mTheta gradient}) with $\mTheta^{(t-1)},\vp^{(t-1)},\mW^{(t-1)}$
    \STATE $\mTheta^{(t)},\vp^{(t)} = \sampler(\nabla_{\vp,\mTheta}U;\Psi)$
    \IF{storing condition met}
        \STATE $\{\vp,\mTheta\}\leftarrow \vp^{(t)},\mTheta^{(t)}$
    \ENDIF
    \STATE Maximize ELBO (\cref{eq: ELBO for W}) w.r.t. $\phi$ with $\vp^{(t)}, \mTheta^{(t)}$
\ENDFOR
\end{algorithmic}
\end{algorithm}
\paragraph{SG-MCMC with continuous relaxation}
Furthermore, we explore an alternative formulation that circumvents the need for variational inference. Instead, we employ SG-MCMC to sample $\tmW$, a continuous relaxation of $\mW$, facilitating a fully sampling-based approach. For a detailed formulation, please refer to \cref{appsec: joint inference SG-MCMC}. We report its performance in \cref{appsubsec: fully SG-MCMC performance}, which surprisingly is inferior to SG-MCMC+VI. We hypothesize that coupling $\mW, \vp$ through $\mu_\phi$ is important since changes in $\vp$ results in changes of the permutation matrix $\perm(\vp)$, which should also influence $\mW$ accordingly during posterior inference. However, through sampling $\tmW$ with few SG-MCMC steps, this change cannot be immediately reflected, resulting in inferior performance. Thus, we focus only on the performance of SG-MCMC+VI for our experiments. 

\paragraph{Computational complexity}
Our proposed SG-MCMC+VI offers a notable improvement in computational cost compared to existing approaches, such as DIBS \cite{lorch2021dibs}. 
The computational complexity of our method is $O(BN_p+N_pd^3)$, where $B$ represents the batch size and $N_p$ is the number of parallel SG-MCMC chains. This former term stems from the forward and backward passes, and the latter comes from the Hungarian algorithm, which can be parallelized to further reduce computational cost. In comparison, DIBS has a complexity of $O(N_p^2N+N_pd^3)$ with $N\gg B$ being the full dataset size. This is due to the kernel computation involving the entire dataset and the evaluation of the matrix exponential in the DAG regularizer \cite{zheng2018dags}. As a result, our approach provides linear scalability w.r.t. $N_p$ with substantially smaller batch size $B$. Conversely, DIBS exhibits quadratic scaling in terms of $N_p$ and lacks support for mini-batch gradients.

\section{Related Work}
\label{appsec: related work}
Bayesian causal discovery literature has primarily focused on inference in linear models with closed-form posteriors or marginalized parameters. Early works considered sampling directed acyclic graphs (DAGs) for discrete~\cite{cooper1992bayesian, madigan1995bayesian, heckerman2006bayesian} and Gaussian random variables~\cite{friedman2003being, tong2001active} using Markov chain Monte Carlo (MCMC) in the DAG space. However, these approaches exhibit slow mixing and convergence~\cite{eaton2012bayesian,grzegorczyk2008improving}, often requiring restrictions on number of parents~\cite{kuipers2017partition}. Alternative exact dynamic programming methods are limited to low-dimensional settings~\cite{koivisto2012advances}. 

Recent advances in variational inference~\cite{zhang2018advances} have facilitated graph inference in DAG space, with gradient-based methods employing the NOTEARS DAG penalty \cite{zheng2018dags}.\cite{annadani2021variational} samples DAGs from autoregressive adjacency matrix distributions, while \cite{lorch2021dibs} utilizes Stein variational approach \cite{liu2016stein} for DAGs and causal model parameters. \cite{cundy2021bcd} proposed a variational inference framework on node orderings using the gumbel-sinkhorn gradient estimator \cite{mena2018learning}. \cite{deleu2022bayesian,nishikawa2022bayesian} employ the GFlowNet framework \cite{bengio2021gflownet} for inferring the DAG posterior. Most methods, except\cite{lorch2021dibs} are restricted to linear models, while \cite{lorch2021dibs} has high computational costs and lacks DAG generation guarantees compared to our method.

In contrast, \emph{quasi-Bayesian} methods, such as DAG bootstrap \cite{friedman2013data}, demonstrate competitive performance. DAG bootstrap resamples data and estimates a single DAG using PC \cite{spirtes2000causation}, GES \cite{chickering2002optimal}, or similar algorithms, weighting the obtained DAGs by their unnormalized posterior probabilities. Recent neural network-based works employ variational inference to learn DAG distributions and point estimates for nonlinear model parameters \cite{charpentier2022differentiable,geffner2022deep}.
\section{Experiments}
\label{sec: experiments}
In this section, we aim to study empirically the following aspects: (1) posterior inference quality of \ModelName~as compared to the true posterior when the causal model is identifiable only upto Markov Equivalence Class (MEC); (2) posterior inference quality of \ModelName~in high dimensional nonlinear causal models (3) ablation studies of \ModelName~ and (4) performance in semi-synthetic and real world applications. The experiment details are included in \cref{app:experiments}.

\paragraph{Baselines.} We mainly compare \ModelName~ with the following baselines: Bootstrap GES (\textbf{BGES})~\cite{chickering2002optimal,friedman2013data}, \textbf{BCD} Nets~\cite{cundy2021bcd}, Differentiable DAG Sampling (\textbf{DDS} \cite{charpentier2022differentiable}) and \textbf{DIBS}~\cite{lorch2021dibs}. 
\begin{wrapfigure}[20]{r}{0.5\textwidth}
    \vspace{2em}
    \centering
    \includegraphics[width=\linewidth]{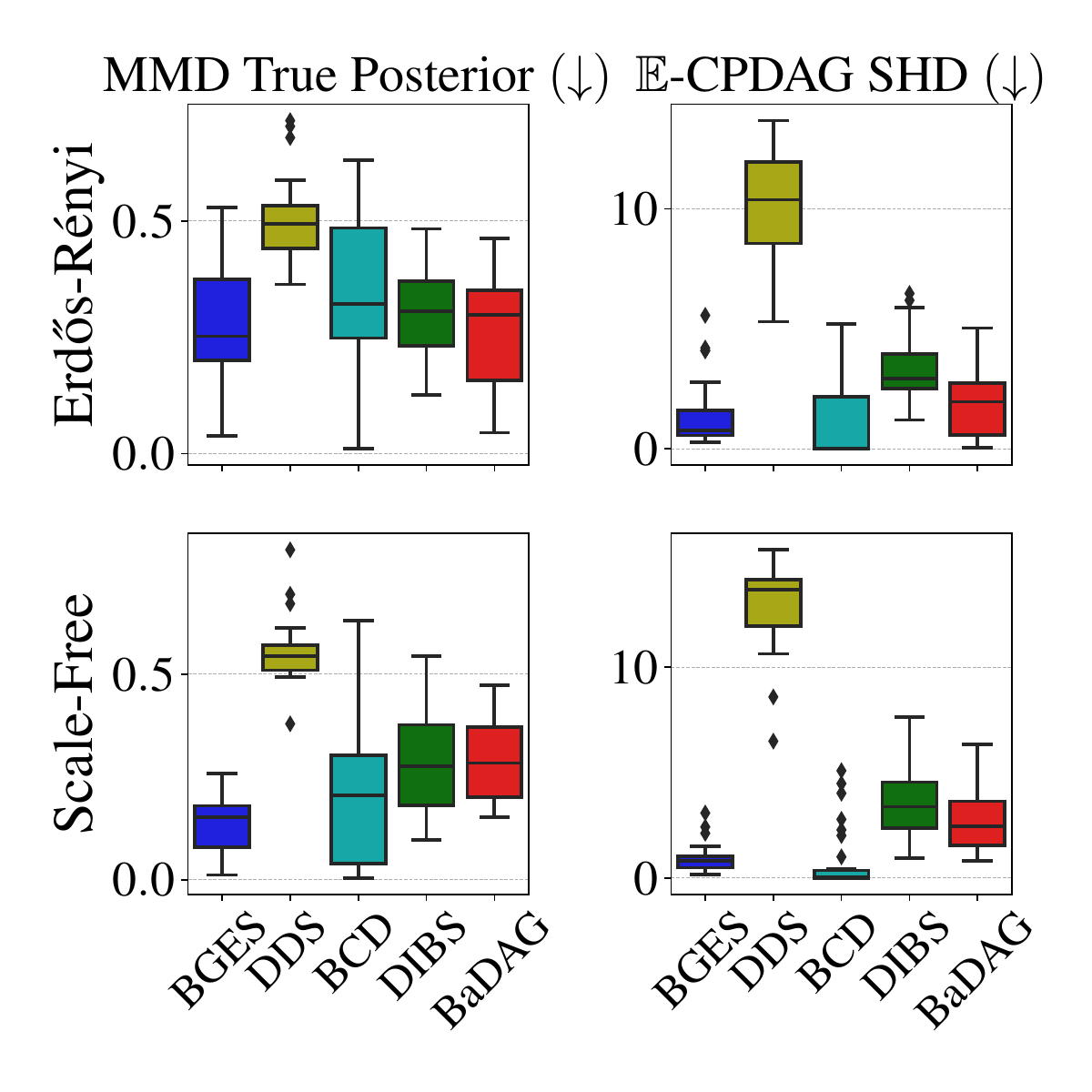}
    \caption{\looseness=-1 Posterior inference on linear synthetic datasets with $d=5$. Metrics are computed against the true posterior. $\downarrow$ denotes lower is better.}
    \label{fig:linear_5_5}
\end{wrapfigure}
\subsection{Evaluation on Synthetic Data}
\paragraph{Synthetic data.}

We evaluate our method on synthetic data, where ground truth graphs are known. Following previous work, we generate data by randomly sampling DAGs from Erdos-Rènyi (ER)~\cite{erdHos1960evolution} or Scale-Free (SF)~\cite{barabasi1999emergence} graphs with per node degree 2 and drawing at random ground truth parameters for linear or nonlinear models. For $d=5$, we use $N=500$ training, while for higher dimensions, we use $N=5000$. We assess performance on 30 random datasets for each setting. 
 
\paragraph{Metrics} For $d=5$ linear models, we compare the approximate and true posterior over DAGs using Maximum Mean Discrepancy (MMD) and also evaluate the expected CPDAG Structural Hamming Distance (SHD). For higher-dimensional nonlinear models with intractable posterior, we compute the expected SHD (\textbf{$\E$-SHD}), expected orientation F1 score (\textbf{Edge F1}) and negative log-likelihood of the held-out data (\textbf{NLL}). Our synthetic data generation and evaluation protocol follows prior work~\cite{annadani2021variational, geffner2022deep,lorch2021dibs}. All the experimental details, including how we use cross-validation to select hyperparameters is in \cref{app:experiments}.

\subsubsection{Comparison with True Posterior}
  
Capturing equivalence classes and quantifying epistemic uncertainty are crucial in Bayesian causal discovery. We benchmark our 
method
against the true posterior using a 5-variable linear SCM with unequal noise variance (identifiable upto MEC~\cite{peters2014identifiability}). The true posterior over graphs $p(\mG\mid \mD)$ can be computed using the BGe score~\cite{geiger2002parameter,kuipers2014addendum}. Results in \cref{fig:linear_5_5} show that our method outperforms
DIBS and DDS in both ER and SF settings. Compared to BCD, we perform better in terms of MMD in ER but worse in SF. We find that BGES performs very well in low-dimensional linear settings, but suffers significantly in more realistic nonlinear settings (see below).
\begin{wraptable}{r}{0.5\textwidth}
\vspace{2em}
\caption{$\E$-SHD (with $95\%$ CI) for ER graphs in higher dimensional nonlinear causal models. DIBS becomes computationally prohibitive for $d>50$.}
\label{tab:scaling_Exp}
\begin{adjustbox}{width=0.5\textwidth}
\begin{tabular}{l|c|c}
      & $d=70$ & $d=100$ \\ \hline
BGES     &  355.77 $\pm$ 18.02   & 563.02 $\pm$ 27.21      \\
BCD    & 217.05 $\pm$ 9.58    &362.66 $\pm$ 29.18       \\
DIBS      &  N/A   &  N/A     \\
BaDAG     & \textbf{143.70 $\pm$ 11.61}    & \textbf{295.92 $\pm$ 24.67}    
\end{tabular}
\end{adjustbox}
\vspace{-3em}
\end{wraptable}
\subsubsection{Evaluation in Higher Dimensions}
\begin{figure}
    \centering
    \includegraphics[width=\linewidth]{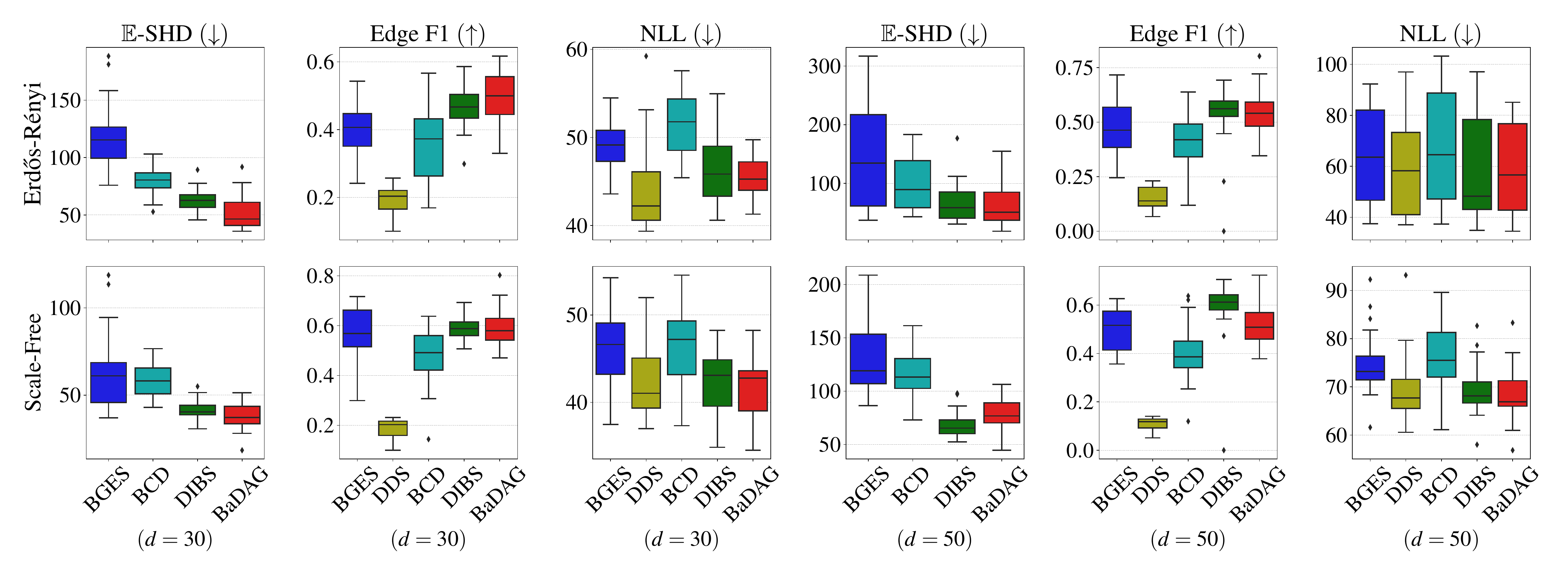}
    \caption{Posterior inference of both graph and functional parameters on synthetic datasets of nonlinear causal models with $d=30$ and $d=50$ variables. \ModelName~gives best results across most metrics and outperforms other permutation based approaches (BCD and DDS). We found DDS to perform significantly worse in terms of $\E$-SHD and thus has been omitted for clarity. $\downarrow$ denotes lower is better and $\uparrow$ denotes higher is better.}
    \label{fig:nonlinear_30_60}
\end{figure}

We evaluate our method on high dimensional scenarios with nonlinear relations. Our approach is the first to attempt full posterior inference in nonlinear models using permutation-based methods. Results for $d=30$ variables in \cref{fig:nonlinear_30_60} demonstrate that \ModelName~ significantly outperforms other \emph{permutation-based approaches} and DIBS in most of the metrics. For $d=50$, \ModelName~ performs comparably to DIBS in ER but a little worse in SF. However, our method achieves better NLL on held-out data compared to most baselines including DIBS for $d=30,50$, ER and SF settings. Only DDS gives better NLL for $d=30$ ER setting, but this doesn't translate well to other metrics and settings. 
We additionally evaluate on $d\in \{70,100\}$ variables (\cref{tab:scaling_Exp}). We find that our method consistently outperforms the baselines with $d=70$ and in terms of $\mathbb{E}$-SHD with $d=100$. Full results are presented in \cref{appsubsec: higher dimensional datasets}. Competitive performance for $d>50$ in nonlinear settings further demonstrates the applicability and computational efficiency of the proposed approach. In contrast, the only fully Bayesian nonlinear method, DIBS, is not computationally efficient to run for $d>50$.

\subsection{Ablation Studies}
\begin{figure}
\centering
\begin{subfigure}{.48\textwidth}
    \centering
    \includegraphics[width=\linewidth]{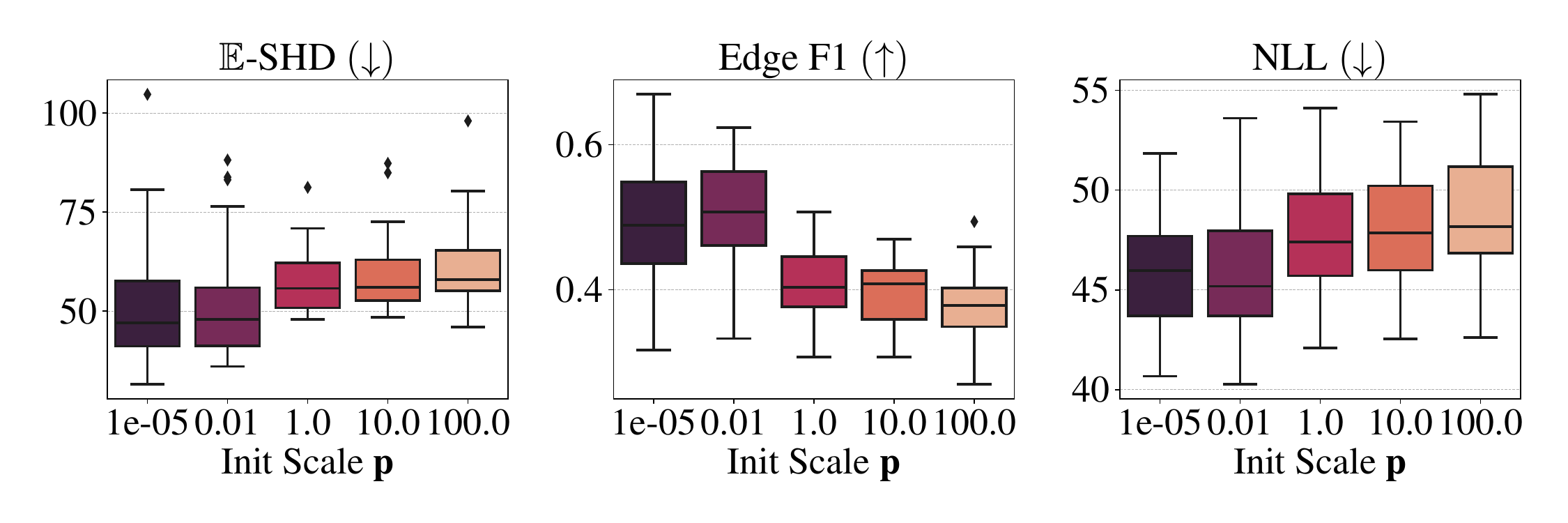}
    \caption{Posterior inference with different initialized $\vp$ scale. }\label{fig: ablation init p scale}
\end{subfigure}
    \hfill
\begin{subfigure}{.48\textwidth}
    \centering
    \includegraphics[width=\linewidth]{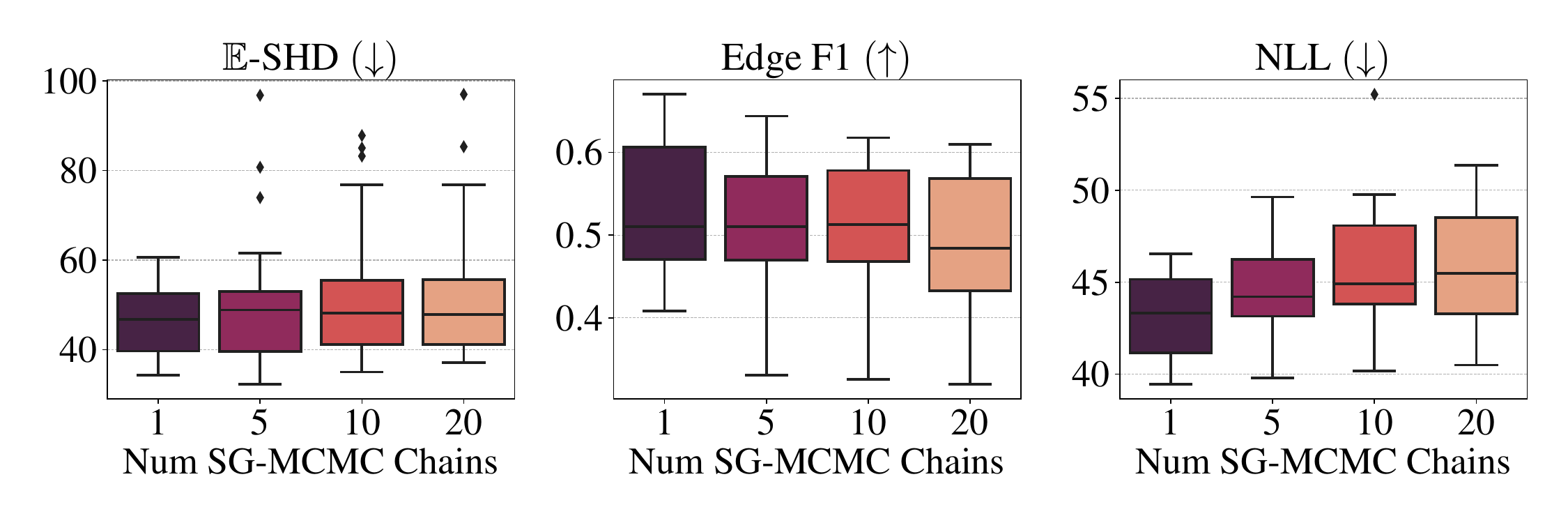}
    \caption{Posterior inference with different number of parallel SG-MCMC chains. }\label{fig: ablation num particles}
\end{subfigure}
   \bigskip
\begin{subfigure}{.48\textwidth}
    \centering
    \includegraphics[width=\linewidth]{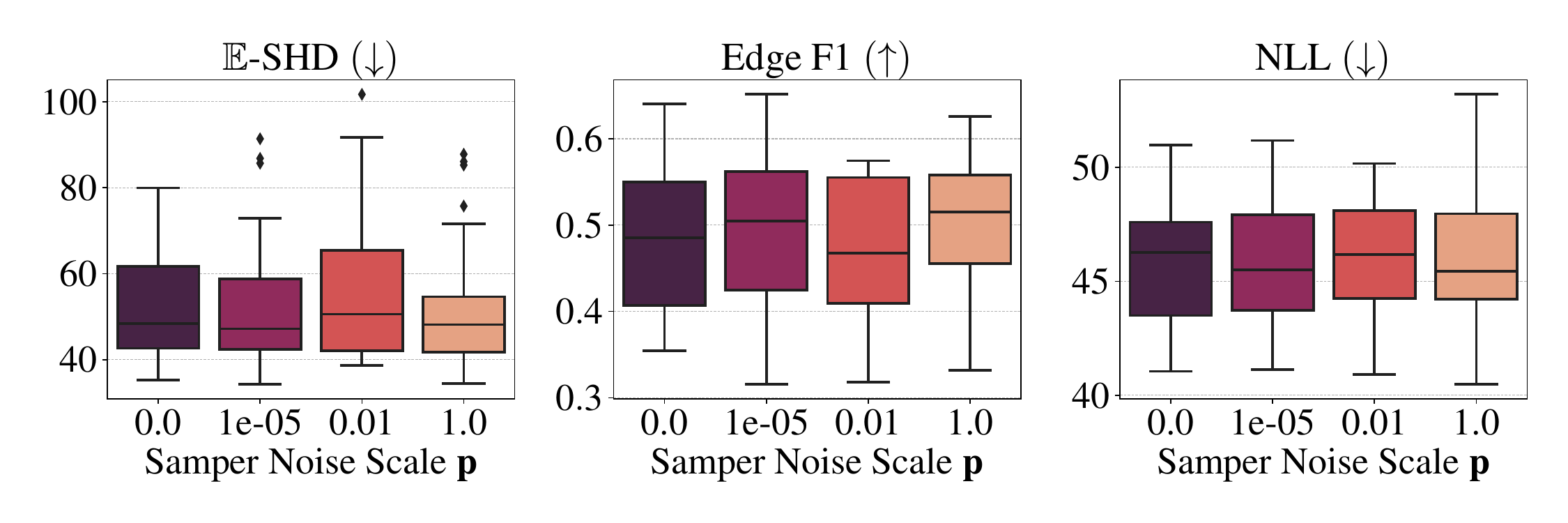}
    \caption{Posterior inference with different level of injected noise scale for $\vp$.}\label{fig: ablation scale_p}
\end{subfigure}
\hfill
\begin{subfigure}{0.48\textwidth}
  \centering
  \includegraphics[width=\linewidth]{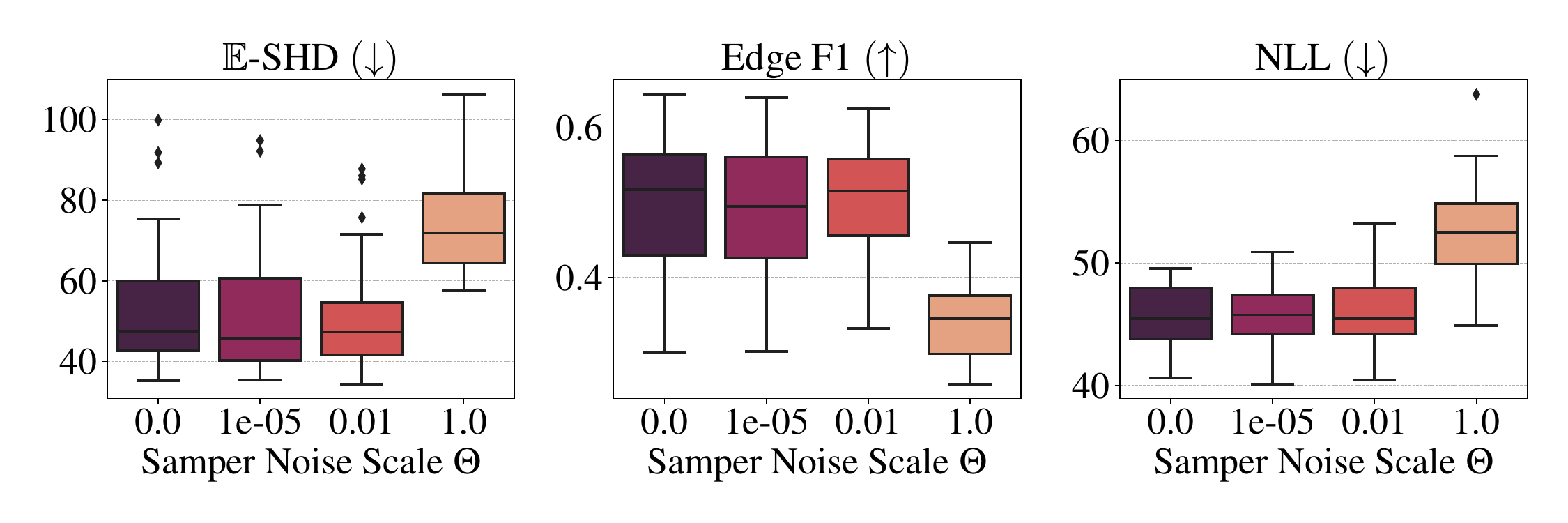}
  \caption{Posterior inference  with different level of injected noise scale for $\mTheta$. }\label{fig: ablation scale}
\end{subfigure} 
\caption{Ablation study of posterior inference quality of \ModelName~on $d=30$ ER synthetic dataset.}
\label{fig:images}
\end{figure}

We conduct ablation studies on our method using the nonlinear ER $d=30$ dataset. 
\paragraph{Initialized $\vp$ scale} \cref{fig: ablation init p scale} investigates the influence of the initialized scale of $\vp$. We found that the performance is the best with $\alpha=0.01$ or $10^{-5}$, and deteriorates with increasing scales. This is because with larger initialization scale, the absolute value of the $\vp$ is large. Longer SG-MCMC updates are needed to reverse the node potential order, which hinders the exploration of possible permutations, resulting in the convergence to poor local optima.

\paragraph{Number of SG-MCMC chains}
We examine the impact of the number of parallel SG-MCMC chains in \cref{fig: ablation num particles}. We observe that it does not have a significant impact on the performance, especially with respect to the $\E$-SHD and Edge F1 metrics.
\paragraph{Injected noise level for SG-MCMC}
In \cref{fig: ablation scale,fig: ablation scale_p}, we study the performance differences arising from various injected noise levels for $\vp$ and $\mTheta$ in the SG-MCMC algorithm (i.e.~$s$ of the SG-MCMC formulation in \cref{app: SG-MCMC update}). Interestingly, the noise level of $\vp$ does not impact the performance as much as the level of $\mTheta$. Injecting noise helps improve the performance, but a smaller noise level should be chosen for $\mTheta$ to avoid divergence from optima.

\subsection{Application 1: Evaluation on Semi-Synthetic Data}
We evaluate our method on the SynTReN simulator~\cite{van2006syntren}. This simulator creates synthetic transcriptional regulatory
networks and produces simulated gene expression data that approximates real experimental data. We use five different simulated datasets provided by \cite{lachapelle2019gradient} with $N=500$ samples each. \cref{tab:real_results} presents the results of all the methods. We find that our method recovers the true network much better in terms of $\mathbb{E}$-SHD as well as Edge F1 compared to baselines.

\subsection{Application 2: Evaluation on Real Data}
We also evaluate on a real dataset which measures the expression
level of different proteins and phospholipids in human cells (called the Sachs Protein Cells Dataset)~\cite{sachs2005causal}. The data corresponds to a network of protein-protein interactions of 11 different proteins with 17 edges in total among them. There are 853 observational samples in total, from which we bootstrap 800 samples of $5$ different datasets. It is to be noted that this data does not necessarily adhere to the additive noise and DAG assumptions, thereby having significant model misspecification. Results in \cref{tab:real_results} demonstrate that our method performs well as compared to the baselines even with model misspecification, proving the suitability of the proposed framework for real-world settings.

\begin{table}[]
\centering
\caption{Results (with 95$\%$ confidence intervals) on Syntren (semi-synthetic) and Sachs Protein Cells (real-world) datasets. For Syntren, results are averaged over 5 different datasets. For Sachs, results are averaged over 5 different restarts. $\downarrow$ denotes lower is better and $\uparrow$ denotes higher is better. }
\label{tab:real_results}
\begin{tabular}{lcc||cc}
\hline
                           & \multicolumn{2}{c||}{Syntren $(d=20)$}                                                                                 & \multicolumn{2}{c}{Sachs Protein Cells $(d=11)$}                                                                      \\ \hline
                           & \multicolumn{1}{c|}{$\mathbb{E}$-SHD $(\downarrow)$} & Edge F1 $(\uparrow)$ & \multicolumn{1}{c|}{$\mathbb{E}$-SHD $(\downarrow)$} & Edge F1 $(\uparrow)$ \\ \hline
\multicolumn{1}{l|}{BGES}  & \multicolumn{1}{c|}{66.18 $\pm$ 9.47}                & \textbf{0.21 $\pm$ 0.05}    & \multicolumn{1}{c|}{\textbf{16.61 $\pm$ 0.44}}                & 0.22 $\pm$ 0.02      \\
\multicolumn{1}{l|}{DDS}   & \multicolumn{1}{c|}{134.37 $\pm$ 4.58}                                &   0.13 $\pm$ 0.02                  &   \multicolumn{1}{c|}{34.90 $\pm$ 0.73}                                               &  0.21 $\pm$ 0.02                              \\
\multicolumn{1}{l|}{BCD}   & \multicolumn{1}{c|}{38.38 $\pm$ 7.12}                & 0.15 $\pm$ 0.07      & \multicolumn{1}{c|}{17.05 $\pm$ 1.93}                & 0.20 $\pm$ 0.08      \\
\multicolumn{1}{l|}{DIBS}  & \multicolumn{1}{c|}{46.43 $\pm$ 4.12}                & 0.16 $\pm$ 0.02      & \multicolumn{1}{c|}{22.3 $\pm$ 0.31}                 & 0.20 $\pm$ 0.01        \\
\multicolumn{1}{l|}{BaDAG} & \multicolumn{1}{c|}{\textbf{34.21 $\pm$ 2.82}}                                & \textbf{0.20 $\pm$ 0.02}                                    & \multicolumn{1}{c|}{18.92 $\pm$ 1.0}                              & \textbf{0.26 $\pm$ 0.04}                \\
\hline
\end{tabular}
\end{table}

\section{Discussion}
\label{sec: discussion}

In this work, we propose \ModelName, a novel, scalable Bayesian causal discovery framework that employs SG-MCMC (and VI) to infer causal models. We establish the validity of performing Bayesian inference in the augmented $(\mW,\vp)$ space and demonstrate its connection to permutation-based DAG learning. Furthermore, we provide two instantiations of the proposed framework that offers direct DAG sampling and model-agnosticism to linear and nonlinear relations. We demonstrate superior inference accuracy and scalability on various datasets. Future work can address some limitations: (1) designing better variational networks $\mu_\phi$ to capture the complex distributions of $\mW$ compared to the simple independent Bernoulli distribution; (2) improving the performance of SG-MCMC with continuous relaxation (\cref{appsec: joint inference SG-MCMC}), which currently does not align with its theoretical advantages compared to the SG-MCMC+VI counterpart.

\paragraph{Acknowledgements.} The authors would like to thank Colleen Tyler, Maria Defante, and Lisa Parks for conversations on real-world use cases that inspired this work. YA and SB are thankful for the Swedish National Computing's Berzelius cluster for providing resources that were helpful in running some of the baselines of the paper. In addition, the authors would like to thank the anonymous reviewers for their feedback.

\bibliography{reference}

\begin{thebibliography}{10}

\bibitem{adams2011ranking}
Ryan~Prescott Adams and Richard~S Zemel.
\newblock Ranking via sinkhorn propagation.
\newblock {\em arXiv preprint arXiv:1106.1925}, 2011.

\bibitem{agrawal2019abcd}
Raj Agrawal, Chandler Squires, Karren Yang, Karthikeyan Shanmugam, and Caroline
  Uhler.
\newblock Abcd-strategy: Budgeted experimental design for targeted causal
  structure discovery.
\newblock In {\em The 22nd International Conference on Artificial Intelligence
  and Statistics}, pages 3400--3409. PMLR, 2019.

\bibitem{annadani2021variational}
Yashas Annadani, Jonas Rothfuss, Alexandre Lacoste, Nino Scherrer, Anirudh
  Goyal, Yoshua Bengio, and Stefan Bauer.
\newblock Variational causal networks: Approximate bayesian inference over
  causal structures.
\newblock {\em arXiv preprint arXiv:2106.07635}, 2021.

\bibitem{annadani2023differentiable}
Yashas Annadani, Panagiotis Tigas, Desi~R Ivanova, Andrew Jesson, Yarin Gal,
  Adam Foster, and Stefan Bauer.
\newblock Differentiable multi-target causal bayesian experimental design.
\newblock {\em arXiv preprint arXiv:2302.10607}, 2023.

\bibitem{barabasi1999emergence}
Albert-L{\'a}szl{\'o} Barab{\'a}si and R{\'e}ka Albert.
\newblock Emergence of scaling in random networks.
\newblock {\em science}, 286(5439):509--512, 1999.

\bibitem{bengio2021gflownet}
Yoshua Bengio, Salem Lahlou, Tristan Deleu, Edward~J Hu, Mo~Tiwari, and
  Emmanuel Bengio.
\newblock Gflownet foundations.
\newblock {\em arXiv preprint arXiv:2111.09266}, 2021.

\bibitem{bengio2013estimating}
Yoshua Bengio, Nicholas L{\'e}onard, and Aaron Courville.
\newblock Estimating or propagating gradients through stochastic neurons for
  conditional computation.
\newblock {\em arXiv preprint arXiv:1308.3432}, 2013.

\bibitem{blondel2020fast}
Mathieu Blondel, Olivier Teboul, Quentin Berthet, and Josip Djolonga.
\newblock Fast differentiable sorting and ranking.
\newblock In {\em International Conference on Machine Learning}, pages
  950--959. PMLR, 2020.

\bibitem{bongers2021foundations}
Stephan Bongers, Patrick Forr{\'e}, Jonas Peters, and Joris~M Mooij.
\newblock Foundations of structural causal models with cycles and latent
  variables.
\newblock {\em The Annals of Statistics}, 49(5):2885--2915, 2021.

\bibitem{casella1992explaining}
George Casella and Edward~I George.
\newblock Explaining the gibbs sampler.
\newblock {\em The American Statistician}, 46(3):167--174, 1992.

\bibitem{charpentier2022differentiable}
Bertrand Charpentier, Simon Kibler, and Stephan G{\"u}nnemann.
\newblock Differentiable dag sampling.
\newblock {\em arXiv preprint arXiv:2203.08509}, 2022.

\bibitem{chen2015convergence}
Changyou Chen, Nan Ding, and Lawrence Carin.
\newblock On the convergence of stochastic gradient mcmc algorithms with
  high-order integrators.
\newblock {\em Advances in neural information processing systems}, 28, 2015.

\bibitem{chen2014stochastic}
Tianqi Chen, Emily Fox, and Carlos Guestrin.
\newblock Stochastic gradient hamiltonian monte carlo.
\newblock In {\em International conference on machine learning}, pages
  1683--1691. PMLR, 2014.

\bibitem{chickering2002optimal}
David~Maxwell Chickering.
\newblock Optimal structure identification with greedy search.
\newblock {\em Journal of machine learning research}, 3(Nov):507--554, 2002.

\bibitem{cooper1992bayesian}
Gregory~F Cooper and Edward Herskovits.
\newblock A bayesian method for the induction of probabilistic networks from
  data.
\newblock {\em Machine learning}, 9:309--347, 1992.

\bibitem{cundy2021bcd}
Chris Cundy, Aditya Grover, and Stefano Ermon.
\newblock Bcd nets: Scalable variational approaches for bayesian causal
  discovery.
\newblock {\em Advances in Neural Information Processing Systems},
  34:7095--7110, 2021.

\bibitem{deleu2022bayesian}
Tristan Deleu, Ant{\'o}nio G{\'o}is, Chris Emezue, Mansi Rankawat, Simon
  Lacoste-Julien, Stefan Bauer, and Yoshua Bengio.
\newblock Bayesian structure learning with generative flow networks.
\newblock In {\em Uncertainty in Artificial Intelligence}, pages 518--528.
  PMLR, 2022.

\bibitem{eaton2012bayesian}
Daniel Eaton and Kevin Murphy.
\newblock Bayesian structure learning using dynamic programming and mcmc.
\newblock {\em arXiv preprint arXiv:1206.5247}, 2012.

\bibitem{erdHos1960evolution}
Paul Erd{\H{o}}s, Alfr{\'e}d R{\'e}nyi, et~al.
\newblock On the evolution of random graphs.
\newblock {\em Publ. Math. Inst. Hung. Acad. Sci}, 5(1):17--60, 1960.

\bibitem{friedman2013data}
Nir Friedman, Moises Goldszmidt, and Abraham Wyner.
\newblock Data analysis with bayesian networks: A bootstrap approach.
\newblock {\em arXiv preprint arXiv:1301.6695}, 2013.

\bibitem{friedman2003being}
Nir Friedman and Daphne Koller.
\newblock Being bayesian about network structure. a bayesian approach to
  structure discovery in bayesian networks.
\newblock {\em Machine learning}, 50(1):95--125, 2003.

\bibitem{geffner2022deep}
Tomas Geffner, Javier Antoran, Adam Foster, Wenbo Gong, Chao Ma, Emre Kiciman,
  Amit Sharma, Angus Lamb, Martin Kukla, Nick Pawlowski, et~al.
\newblock Deep end-to-end causal inference.
\newblock {\em arXiv preprint arXiv:2202.02195}, 2022.

\bibitem{geiger2002parameter}
Dan Geiger and David Heckerman.
\newblock Parameter priors for directed acyclic graphical models and the
  characterization of several probability distributions.
\newblock {\em The Annals of Statistics}, 30(5):1412--1440, 2002.

\bibitem{gong2022advances}
Wenbo Gong.
\newblock {\em Advances in approximate inference: combining VI and MCMC and
  improving on Stein discrepancy}.
\newblock PhD thesis, University of Cambridge, 2022.

\bibitem{gong2022rhino}
Wenbo Gong, Joel Jennings, Cheng Zhang, and Nick Pawlowski.
\newblock Rhino: Deep causal temporal relationship learning with
  history-dependent noise.
\newblock {\em arXiv preprint arXiv:2210.14706}, 2022.

\bibitem{gong2018meta}
Wenbo Gong, Yingzhen Li, and Jos{\'e}~Miguel Hern{\'a}ndez-Lobato.
\newblock Meta-learning for stochastic gradient mcmc.
\newblock {\em arXiv preprint arXiv:1806.04522}, 2018.

\bibitem{gong2020sliced}
Wenbo Gong, Yingzhen Li, and Jos{\'e}~Miguel Hern{\'a}ndez-Lobato.
\newblock Sliced kernelized stein discrepancy.
\newblock {\em arXiv preprint arXiv:2006.16531}, 2020.

\bibitem{gong2019icebreaker}
Wenbo Gong, Sebastian Tschiatschek, Sebastian Nowozin, Richard~E Turner,
  Jos{\'e}~Miguel Hern{\'a}ndez-Lobato, and Cheng Zhang.
\newblock Icebreaker: Element-wise efficient information acquisition with a
  bayesian deep latent gaussian model.
\newblock {\em Advances in neural information processing systems}, 32, 2019.

\bibitem{gong2021active}
Wenbo Gong, Kaibo Zhang, Yingzhen Li, and Jos{\'e}~Miguel Hern{\'a}ndez-Lobato.
\newblock Active slices for sliced stein discrepancy.
\newblock In {\em International Conference on Machine Learning}, pages
  3766--3776. PMLR, 2021.

\bibitem{grathwohl2021oops}
Will Grathwohl, Kevin Swersky, Milad Hashemi, David Duvenaud, and Chris
  Maddison.
\newblock Oops i took a gradient: Scalable sampling for discrete distributions.
\newblock In {\em International Conference on Machine Learning}, pages
  3831--3841. PMLR, 2021.

\bibitem{gretton2012kernel}
Arthur Gretton, Karsten~M Borgwardt, Malte~J Rasch, Bernhard Sch{\"o}lkopf, and
  Alexander Smola.
\newblock A kernel two-sample test.
\newblock {\em The Journal of Machine Learning Research}, 13(1):723--773, 2012.

\bibitem{grzegorczyk2008improving}
Marco Grzegorczyk and Dirk Husmeier.
\newblock Improving the structure mcmc sampler for bayesian networks by
  introducing a new edge reversal move.
\newblock {\em Machine Learning}, 71(2-3):265, 2008.

\bibitem{heckerman2006bayesian}
David Heckerman, Christopher Meek, and Gregory Cooper.
\newblock A bayesian approach to causal discovery.
\newblock {\em Innovations in Machine Learning: Theory and Applications}, pages
  1--28, 2006.

\bibitem{hoyer2008nonlinear}
Patrik Hoyer, Dominik Janzing, Joris~M Mooij, Jonas Peters, and Bernhard
  Sch{\"o}lkopf.
\newblock Nonlinear causal discovery with additive noise models.
\newblock {\em Advances in neural information processing systems}, 21, 2008.

\bibitem{kingma2014adam}
Diederik~P Kingma and Jimmy Ba.
\newblock Adam: A method for stochastic optimization.
\newblock {\em arXiv preprint arXiv:1412.6980}, 2014.

\bibitem{koivisto2012advances}
Mikko Koivisto.
\newblock Advances in exact bayesian structure discovery in bayesian networks.
\newblock {\em arXiv preprint arXiv:1206.6828}, 2012.

\bibitem{kuhn1955hungarian}
Harold~W Kuhn.
\newblock The hungarian method for the assignment problem.
\newblock {\em Naval research logistics quarterly}, 2(1-2):83--97, 1955.

\bibitem{kuipers2017partition}
Jack Kuipers and Giusi Moffa.
\newblock Partition mcmc for inference on acyclic digraphs.
\newblock {\em Journal of the American Statistical Association},
  112(517):282--299, 2017.

\bibitem{kuipers2014addendum}
Jack Kuipers, Giusi Moffa, and David Heckerman.
\newblock Addendum on the scoring of gaussian directed acyclic graphical
  models.
\newblock 2014.

\bibitem{lachapelle2019gradient}
S{\'e}bastien Lachapelle, Philippe Brouillard, Tristan Deleu, and Simon
  Lacoste-Julien.
\newblock Gradient-based neural dag learning.
\newblock {\em arXiv preprint arXiv:1906.02226}, 2019.

\bibitem{li2016preconditioned}
Chunyuan Li, Changyou Chen, David Carlson, and Lawrence Carin.
\newblock Preconditioned stochastic gradient langevin dynamics for deep neural
  networks.
\newblock In {\em Thirtieth AAAI Conference on Artificial Intelligence}, 2016.

\bibitem{liu2016stein}
Qiang Liu and Dilin Wang.
\newblock Stein variational gradient descent: A general purpose bayesian
  inference algorithm.
\newblock {\em Advances in neural information processing systems}, 29, 2016.

\bibitem{lorch2021dibs}
Lars Lorch, Jonas Rothfuss, Bernhard Sch{\"o}lkopf, and Andreas Krause.
\newblock Dibs: Differentiable bayesian structure learning.
\newblock {\em Advances in Neural Information Processing Systems},
  34:24111--24123, 2021.

\bibitem{ma2019variational}
Chao Ma, Yingzhen Li, and Jos{\'e}~Miguel Hern{\'a}ndez-Lobato.
\newblock Variational implicit processes.
\newblock In {\em International Conference on Machine Learning}, pages
  4222--4233. PMLR, 2019.

\bibitem{ma2015complete}
Yi-An Ma, Tianqi Chen, and Emily Fox.
\newblock A complete recipe for stochastic gradient mcmc.
\newblock {\em Advances in neural information processing systems}, 28, 2015.

\bibitem{madigan1995bayesian}
David Madigan, Jeremy York, and Denis Allard.
\newblock Bayesian graphical models for discrete data.
\newblock {\em International Statistical Review/Revue Internationale de
  Statistique}, pages 215--232, 1995.

\bibitem{mena2018learning}
Gonzalo Mena, David Belanger, Scott Linderman, and Jasper Snoek.
\newblock Learning latent permutations with gumbel-sinkhorn networks.
\newblock {\em arXiv preprint arXiv:1802.08665}, 2018.

\bibitem{munkres1957algorithms}
James Munkres.
\newblock Algorithms for the assignment and transportation problems.
\newblock {\em Journal of the society for industrial and applied mathematics},
  5(1):32--38, 1957.

\bibitem{murphy2001active}
Kevin~P Murphy.
\newblock Active learning of causal bayes net structure.
\newblock Technical report, technical report, UC Berkeley, 2001.

\bibitem{niculae2018sparsemap}
Vlad Niculae, Andre Martins, Mathieu Blondel, and Claire Cardie.
\newblock Sparsemap: Differentiable sparse structured inference.
\newblock In {\em International Conference on Machine Learning}, pages
  3799--3808. PMLR, 2018.

\bibitem{nishikawa2022bayesian}
Mizu Nishikawa-Toomey, Tristan Deleu, Jithendaraa Subramanian, Yoshua Bengio,
  and Laurent Charlin.
\newblock Bayesian learning of causal structure and mechanisms with gflownets
  and variational bayes.
\newblock {\em arXiv preprint arXiv:2211.02763}, 2022.

\bibitem{pearl2009causality}
Judea Pearl.
\newblock {\em Causality}.
\newblock Cambridge university press, 2009.

\bibitem{peters2014identifiability}
Jonas Peters and Peter B{\"u}hlmann.
\newblock Identifiability of gaussian structural equation models with equal
  error variances.
\newblock {\em Biometrika}, 101(1):219--228, 2014.

\bibitem{peters2017elements}
Jonas Peters, Dominik Janzing, and Bernhard Sch{\"o}lkopf.
\newblock {\em Elements of causal inference: foundations and learning
  algorithms}.
\newblock The MIT Press, 2017.

\bibitem{peters2014causal}
Jonas Peters, Joris~M Mooij, Dominik Janzing, and Bernhard Sch{\"o}lkopf.
\newblock Causal discovery with continuous additive noise models.
\newblock 2014.

\bibitem{pe2001inferring}
Dana Pe’er, Aviv Regev, Gal Elidan, and Nir Friedman.
\newblock Inferring subnetworks from perturbed expression profiles.
\newblock {\em Bioinformatics}, 17(suppl\_1):S215--S224, 2001.

\bibitem{platen2010numerical}
Eckhard Platen and Nicola Bruti-Liberati.
\newblock {\em Numerical solution of stochastic differential equations with
  jumps in finance}, volume~64.
\newblock Springer Science \& Business Media, 2010.

\bibitem{robinson1973counting}
Robert~W Robinson.
\newblock Counting labeled acyclic digraphs.
\newblock {\em New directions in the theory of graphs}, pages 239--273, 1973.

\bibitem{sachs2005causal}
Karen Sachs, Omar Perez, Dana Pe'er, Douglas~A Lauffenburger, and Garry~P
  Nolan.
\newblock Causal protein-signaling networks derived from multiparameter
  single-cell data.
\newblock {\em Science}, 308(5721):523--529, 2005.

\bibitem{spirtes2000causation}
Peter Spirtes, Clark~N Glymour, Richard Scheines, and David Heckerman.
\newblock {\em Causation, prediction, and search}.
\newblock MIT press, 2000.

\bibitem{springenberg2016bayesian}
Jost~Tobias Springenberg, Aaron Klein, Stefan Falkner, and Frank Hutter.
\newblock Bayesian optimization with robust bayesian neural networks.
\newblock {\em Advances in neural information processing systems}, 29, 2016.

\bibitem{sun2022discrete}
Haoran Sun, Hanjun Dai, Bo~Dai, Haomin Zhou, and Dale Schuurmans.
\newblock Discrete langevin sampler via wasserstein gradient flow.
\newblock {\em arXiv preprint arXiv:2206.14897}, 2022.

\bibitem{sun2019functional}
Shengyang Sun, Guodong Zhang, Jiaxin Shi, and Roger Grosse.
\newblock Functional variational bayesian neural networks.
\newblock {\em arXiv preprint arXiv:1903.05779}, 2019.

\bibitem{tigas2022interventions}
Panagiotis Tigas, Yashas Annadani, Andrew Jesson, Bernhard Sch{\"o}lkopf, Yarin
  Gal, and Stefan Bauer.
\newblock Interventions, where and how? experimental design for causal models
  at scale.
\newblock {\em Advances in neural information processing systems}, 36, 2022.

\bibitem{tong2001active}
Simon Tong and Daphne Koller.
\newblock Active learning for structure in bayesian networks.
\newblock In {\em International joint conference on artificial intelligence},
  volume~17, pages 863--869. Citeseer, 2001.

\bibitem{trippe2018overpruning}
Brian Trippe and Richard Turner.
\newblock Overpruning in variational bayesian neural networks.
\newblock {\em arXiv preprint arXiv:1801.06230}, 2018.

\bibitem{van2006syntren}
Tim Van~den Bulcke, Koenraad Van~Leemput, Bart Naudts, Piet van Remortel,
  Hongwu Ma, Alain Verschoren, Bart De~Moor, and Kathleen Marchal.
\newblock Syntren: a generator of synthetic gene expression data for design and
  analysis of structure learning algorithms.
\newblock {\em BMC bioinformatics}, 7:1--12, 2006.

\bibitem{van2006application}
Chikako Van~Koten and AR~Gray.
\newblock An application of bayesian network for predicting object-oriented
  software maintainability.
\newblock {\em Information and Software Technology}, 48(1):59--67, 2006.

\bibitem{welling2011bayesian}
Max Welling and Yee~W Teh.
\newblock Bayesian learning via stochastic gradient langevin dynamics.
\newblock In {\em Proceedings of the 28th international conference on machine
  learning (ICML-11)}, pages 681--688, 2011.

\bibitem{ye2017langevin}
Nanyang Ye, Zhanxing Zhu, and Rafal~K Mantiuk.
\newblock Langevin dynamics with continuous tempering for training deep neural
  networks.
\newblock {\em arXiv preprint arXiv:1703.04379}, 2017.

\bibitem{yu2019dag}
Yue Yu, Jie Chen, Tian Gao, and Mo~Yu.
\newblock Dag-gnn: Dag structure learning with graph neural networks.
\newblock In {\em International Conference on Machine Learning}, pages
  7154--7163. PMLR, 2019.

\bibitem{yu2021dags}
Yue Yu, Tian Gao, Naiyu Yin, and Qiang Ji.
\newblock Dags with no curl: An efficient dag structure learning approach.
\newblock In {\em International Conference on Machine Learning}, pages
  12156--12166. PMLR, 2021.

\bibitem{zanella2020informed}
Giacomo Zanella.
\newblock Informed proposals for local mcmc in discrete spaces.
\newblock {\em Journal of the American Statistical Association},
  115(530):852--865, 2020.

\bibitem{zantedeschi2023dag}
Valentina Zantedeschi, Luca Franceschi, Jean Kaddour, Matt~J Kusner, and Vlad
  Niculae.
\newblock Dag learning on the permutahedron.
\newblock {\em arXiv preprint arXiv:2301.11898}, 2023.

\bibitem{zhang2018advances}
Cheng Zhang, Judith B{\"u}tepage, Hedvig Kjellstr{\"o}m, and Stephan Mandt.
\newblock Advances in variational inference.
\newblock {\em IEEE transactions on pattern analysis and machine intelligence},
  41(8):2008--2026, 2018.

\bibitem{zhang2022langevin}
Ruqi Zhang, Xingchao Liu, and Qiang Liu.
\newblock A langevin-like sampler for discrete distributions.
\newblock In {\em International Conference on Machine Learning}, pages
  26375--26396. PMLR, 2022.

\bibitem{zheng2018dags}
Xun Zheng, Bryon Aragam, Pradeep~K Ravikumar, and Eric~P Xing.
\newblock Dags with no tears: Continuous optimization for structure learning.
\newblock {\em Advances in Neural Information Processing Systems}, 31, 2018.

\end{thebibliography}
\bibliographystyle{plain}

\newpage
\appendix
\section*{Appendix -- BayesDAG: Gradient-Based Posterior Inference for Causal Discovery}
\section{Joint Inference with SG-MCMC}
\label{appsec: joint inference SG-MCMC}
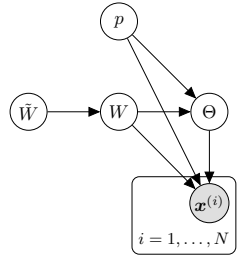
\begin{wrapfigure}[12]{r}{0.4\textwidth}
   \usetikzlibrary{bayesnet}
\scalebox{0.7}{
\begin{tikzpicture}[scale=3]

  \node[latent] (Wtilde) {$\tilde{W}$};
  \node[latent, right=of Wtilde] (W) {$W$};
  \node[latent, above=of W] (p) {$p$};
  \node[latent, right=of W] (Theta) {$\Theta$};
  \node[obs, below=of Theta] (x) {$\vx^{(i)}$};
  
  \edge {Wtilde} {W};
  \edge {p} {x};
  \edge {W} {Theta};
  \edge {W} {x};
  \edge {Theta} {x};
  \edge {p} {Theta};
  
  \plate {x_plate} {(x)} {$i = 1, \dots, N$};

\end{tikzpicture}
}
   \caption{Graphical model with latent variable $\tmW$.}
   \label{fig: graphical model with latent}
\end{wrapfigure}

In this section, we propose an alternative formulation that enables a joint inference framework for $\vp,\mW,\mTheta$ using SG-MCMC, thereby avoiding the need for variational inference for $\mW$.

We adopt a continuous relaxation of $\mW$, similar to \cite{lorch2021dibs}, by introducing a latent variable $\tmW$. The graphical model is illustrated in \cref{fig: graphical model with latent}.  We can define 
\begin{equation}
   p(\mW|\tmW) = \prod_{i,j} p(W_{ij}|\tilde{W}_{ij})
   \label{eq: W and tilde W forward prob}
\end{equation}
with $p(W_{ij}=1|\tilde{W}_{ij})=\sigma(\tilde{W}_{ij})$ where $\sigma(\cdot)$ is the sigmoid function. In other words, $\tilde{W}{ij}$ defines the existence logits of $W{ij}$.

With the introduction of $\tmW$, the original posterior expectations of $p(\vp,\mW,\mTheta\vert \mD)$, e.g.~during evaluation, can be translated using the following proposition. 
\begin{proposition}[Equivalence of posterior expectation]
   Under the generative model \cref{fig: graphical model with latent}, we have
   \begin{equation}
       \E_{p(\vp,\mW,\mTheta|\mD)}\left[
       \vf(\mG=\tau(\vp,\mW),\mTheta)
       \right] = \E_{p(\vp,\tmW,\mTheta)}\left[
       \frac{
       \E_{p(\mW\vert \tmW)}\left[f(\mG,\mTheta)p(\mD,\mTheta\vert \vp,\mW)\right]
       }{
       \E_{p(\mW\vert \tmW)}\left[p(\mD,\mTheta\vert \vp, \mW)\right]
       }
       \right]
       \label{eq: evaluation equivalence}
   \end{equation}
where $\vf$ is the target quantity.
\label{prop: equivalence of evaluation}
\end{proposition}

This proof is in \cref{subapp: proof of prop equivalence evaluation}.

With this proposition, instead of sampling $\mW$, use SG-MCMC to draw $\tmW$ samples. 
Similar to \cref{subsubsec: combined inference}, to use SG-MCMC for $\vp,\tmW,\mTheta$, we need their gradient information. The following proposition specifies the required gradients.
\begin{proposition}
   With the generative model defined as \cref{fig: graphical model with latent}, we have
   \begin{align}
       \nabla_{\vp,\mTheta,\tmW} U(\vp,\tmW,\mTheta) = &-\nabla_{\vp} \log p(\vp) - \nabla_{\mTheta}\log p(\mTheta) \nonumber \\
       &-\nabla_{\tmW}\log p(\tmW) - \nabla_{\vp,\mTheta,\tmW} \log \E_{p(\mW\vert \tmW)}[p(\mD\vert \mW,\vp,\mTheta)]
       \label{eq: joint inference gradient}
   \end{align}
   \label{prop: joint inference gradient}
\end{proposition}
The proof is in \cref{subapp: joint inference gradient}.

With these gradients, we can directly plug in existing SG-MCMC samplers to draw samples for $\vp, \tmW$, and $\mTheta$ in joint inference (\cref{alg: joint inference}). 

\begin{algorithm}[tb]
\caption{Joint inference}
\label{alg: joint inference}
\begin{algorithmic}
\STATE {\bfseries Input:} dataset $\mD$, prior $p(\vp,\tmW,\mTheta)$, SG-MCMC sampler update $\sampler(\cdot)$; sampler hyperparameter $\Psi$; training steps $T$.
\STATE {\bfseries Output:} posterior samples $\{\vp,\tmW,\mTheta\}$
\STATE Initialize $\vp_0,\tmW_0, \mTheta_0$
\FOR{$t=1,\ldots, T$}
   \STATE Evaluate gradient $\nabla_{\vp_{t-1},\tmW_{t-1},\mTheta_{t-1}} U$ based on \cref{eq: joint inference gradient}.
   \STATE Update samples $\vp_t,\tmW_t,\mTheta_t = \sampler(\nabla_{\vp_{t-1},\tmW_{t-1},\mTheta_{t-1}} U; \Psi)$
   \IF{storing condition met}
       \STATE $\{\vp,\tmW,\mTheta\}\leftarrow \vp_t,\tmW_t, \mTheta_t$
   \ENDIF
\ENDFOR
\end{algorithmic}
\end{algorithm}
\section{Theory}
\label{app: Theory}
\subsection{Proof of \cref{thm: equivalence of bayesian inference}}
\label{subapp: proof of equivalence of bayesian inference}
For completeness, we recite the theorem here.
\begin{reptheorem}{thm: equivalence of bayesian inference}[Equivalence of inference in $(\mW,\vp)$ and binary DAG space]
Assume graph $\mG$ is a binary adjacency matrix representing a DAG and node potential $\vp$ does not contain the same values, i.e.~$p_i\neq p_j$ $\forall i,j$. Then, with the induced joint observational distribution $p(\mD,\mG)$, dataset $\mD$ and a corresponding prior $p(\mG)$, we have
\begin{align}
    p(\mG\vert \mD) = \int p_\tau(\vp,\mW\vert \mD)\indicator(\mG=\tau(\mW,\vp))d\mW d\vp
    \label{eq: equivalence of bayesian inference}
\end{align}
if $p(\mG)=\int p_\tau(\vp,\mW)\indicator(\mG=\tau(\mW,\vp))d\mW d\vp$, where $p_\tau(\mW,\vp)$ is the prior, $\indicator(\cdot)$ is the indicator function and $p_\tau(\vp,\mW\vert D)$ is the posterior distribution over $\vp,\mW$. 
\end{reptheorem}

To prove this theorem, we first prove the following lemma stating the equivalence of $\tau$ (\cref{eq: Binary NoCurl}) to binary DAG space. 
\begin{lemma}[Equivalence of $\tau$ to DAG space]
Consider $d$ random variables, a node potential vector $\vp\in\sR^d$ and a binary matrix $\mW\in\binaryset$. Then the following holds:
\begin{enumerate}[(a)]
\item For any $\mW\in\binaryset$, $\vp\in\sR^d$, $\mG=\tau(\mW,\vp)$ is a DAG.
    \item For any DAG $\mG\in \sD$, where $\sD$ is the space of all DAGs, there exists a corresponding $(\mW,\vp)$ such that $\tau(\mW,\vp)=\mG$.
\end{enumerate}
\label{lemma: Equivalence of DAG space}
\end{lemma}
\begin{proof}
The main proof directly follows the theorem 2.1 in \cite{yu2021dags}. For (a), we show the output from $\tau(\mW,\vp)$ must be a DAG. 
By leveraging the Lemma 3.4 in \cite{yu2021dags}, we can easily obtain that $\step(\grad \vp)$ emits a binary adjacency matrix representing a DAG. The only difference is that we replace the $\relu(\cdot)$ with $\step(\cdot)$ but the conclusion can be directly generalized.

For (b), we show that for any DAG $\mG$, there exists a $(\mW,\vp)$ pair s.t. $\tau(\mW,\vp)=\mG$. To see this, we can observe that $\vp$ implicitly defines a topological order in the mapping $\tau$. For any $p_i>p_j$, we have $j\rightarrow i$ after the mapping $\step(\grad \vp)$. Thus, by leveraging Theorem 3.7 in \cite{yu2021dags}, we obtain that there exists a potential vector $\vp\in\sR^d$ for any DAG $\mG$ such that 
$$
(\grad\vp)(i,j)>0 \;\;\;\;\;\text{when}\;G_{ij}=1
$$
Thus, we can choose $\mW$ in the following way:
$$
\mW=\begin{cases}
&W_{ij}=0\;\;\;\;\; \text{if}\; G_{ij}=0\\
&W_{ij}=1\;\;\;\;\; \text{if}\; G_{ij}=1
\end{cases}
$$
\end{proof}
Next, let's prove the \cref{thm: equivalence of bayesian inference}.

\begin{proof}[Proof of \cref{thm: equivalence of bayesian inference}]
    From \cref{lemma: Equivalence of DAG space}, we see that the mapping is complete. Namely, the $(\mW,\vp)$ space can represent the entire DAG space. Next, we show that performing Bayesian inference in $(\mW,\vp)$ space can also correspond to the inference in DAG space. 

    Assume we have the prior $p_\tau(\mW,\vp)$. Then through mapping $\tau$, we implicitly define a prior over the DAG $\mG$ in the following:
    \begin{equation}
        p_\tau(\mG) = \int p_\tau(\mW,\vp)\indicator(\mG=\tau(\mW,\vp))d\mW d\vp
        \label{eq: accumulated prior}
    \end{equation}
    This basically states that the corresponding prior over $\mG$ is an accumulation of the corresponding probability associated with $(\mW,\vp)$ pairs.

    Similarly, we can define a corresponding posterior $p_\tau(\mG\vert \mD)$:
    \begin{equation}
        p_\tau(\mG\vert \mD) = \int p_\tau(\mW,\vp\vert \mD)\indicator(\mG=\tau(\mW,\vp))d\mW d\vp
        \label{eq: accumulated posterior}
    \end{equation}
    Now, let's show that this posterior $p_\tau(\mG\vert \mD) = p(\mG\vert \mD)$ if prior matches, i.e.~$p(\mG)=p_\tau(\mG)$.
    From Bayes's rule, we can easily write down
    \begin{equation}
        p_\tau(\mW,\vp\vert \mD) = \frac{p(\mD\vert \mG=\tau(\mW,\vp))p(\vp,\mW)}{\sum_{\mG'\in\sD}p(\mD,\mG')}
        \label{eq:def of W p posterior}
    \end{equation}
    Then, by substituting \cref{eq:def of W p posterior} into \cref{eq: accumulated posterior}, we have
    \begin{align}
        p_\tau(\mG\vert \mD) &= \int \frac{p(\mD\vert \mG)p_\tau(\mW,\vp)}{\sum_{\mG'\in\sD}p(\mD,\mG')}\indicator(\mG=\tau(\mW,\vp))d\mW d\vp\nonumber \\
        &=\frac{\int p(\mD\vert \mG)p_\tau(\mW,\vp)\indicator(\mG=\tau)d\mW d\vp}{\sum_{\mG'\in\sD}p(\mD,\mG')} \label{eq: proof 3.1 eq 1}\\
        &=\frac{p(\mD\vert \mG)\int p_\tau(\mW,\vp)\indicator(\mG=\tau)d\mW d\vp}{\sum_{\mG'\in\sD}p(\mD,\mG')}\label{eq: proof 3.1 eq 2}\\
        &=\frac{p(\mD\vert \mG)p_\tau(\mG)}{\sum_{\mG'\in\sD}p(\mD\vert\mG')p_\tau(\mG')}\nonumber\\
        &=p(\mG\vert \mD) \label{eq: proof 3.1 eq 3}
    \end{align}
where \cref{eq: proof 3.1 eq 1} is from the fact that $\sum_{\mG'\in\sD}p(\mD,\mG')$ is independent of $(\mW,\vp)$ due to marginalization. \cref{eq: proof 3.1 eq 2} is obtained because $p(\mD\vert \mG)$ is also independent of $(\mW,\vp)$ due to (1) $\indicator(\mG=\tau(\mW,\vp))$ and (2) $p(\mD\vert \mG)$ is a constant when fixing $\mG$. \cref{eq: proof 3.1 eq 3} is obtained by applying Bayes's rule and $p_\tau(\mG) = p(\mG)$. 
\end{proof}

\subsection{Proof of \cref{thm: equivalence of alternative formulation}}
\label{subapp: proof of theorem alternative formulation}

\begin{reptheorem}{thm: equivalence of alternative formulation}[Equivalence of NoCurl formulation]
Assuming the conditions in \cref{thm: equivalence of bayesian inference} are satisfied. Then, for a given $(\mW,\vp)$, we have
$$
\mG=\mW\odot \step(\grad \vp) = \mW\odot \left[ \perm^*(\vp)\mL\perm^*(\vp)^T\right]
$$
where $\mG$ is a DAG and $\perm^*(\vp)$ is defined in \cref{eq: alternative formulation}. 
\end{reptheorem}

To prove this theorem, we need to first prove the following lemma. 

\begin{lemma}
    For any permutation matrix $\mM\in\bm{\Sigma}_d$, we have
    $$
    \grad(\mM \vp) = \mM^T\grad(p) \mM
    $$
    where $\grad$ is the operator defined in \cref{eq: grad operator}.
    \label{lemma: grad with permutation matrix}
\end{lemma}
\begin{proof}
    By definition of $\grad(\cdot)$, we have
    \begin{align*}
        \grad(\mM\vp) &= (\mM\vp)_i - (\mM\vp)_j\\
        &=\bm{1}(i)^T\mM\vp - \bm{1}(j)^T\mM\vp\\
        &=\mM_{i,:}\vp - \mM_{j,:}\vp\\
    \end{align*}
    where $\bm{1}(i)$ is a one-hot vector with $i^{\text{th}}$ entry $1$, and $\mM_{i,:}$ is the $i^{\text{th}}$ row of matrix $\mM$. The above is equivalent to computing the $\grad$ with new labels obtained by permuting $\vp$ with $\mM$. Therefore, we can see that $\grad(\mM\vp)$ can be computed by permuting the original $\grad(\vp)$ by matrix $\mM$.
    $$
    \grad(\mM\vp)=\mM^T\grad(\vp)\mM
    $$
\end{proof}

\begin{proof}[Proof of \cref{thm: equivalence of alternative formulation}]
Since $\mW$ plays the same role in both formulations, we focus on the equivalence of $\step(\grad(\cdot))$.

Define a sorted $\tilde{\vp}= \perm \vp$, where $\perm\in\bm{\Sigma}_d$, such that for $i<j$, we have $\tilde{p}_i > \tilde{p}_j$. Namely, $\perm$ is a permutation matrix.
Thus, we have 
$$
\grad(\vp) = \grad(\perm^T\tilde{\vp}).
$$
By \cref{lemma: grad with permutation matrix}, we have
$$
\grad(\perm^T\tilde{\vp}) = \perm\grad(\tilde{\vp}) \perm^T.
$$
Since $\tilde{\vp}$ is an ordered vector. Therefore, $\grad(\tilde{\vp})$ is a skew-symmetric matrix with a positive lower half part. 

Therefore, we have
$$
\step(\grad(\vp)) = \step(\perm\grad(\tilde{\vp})\perm^T) = \perm\step(\grad(\tilde{p}))\perm^T =  \perm\mL\perm^T
$$
This is true because $\perm$ is just a permutation matrix that does not alter the sign of $\grad(\tilde{\vp})$.

Since $\perm$ is a permutation matrix that sort $\vp$ value in a ascending order, from Lemma 1 in \cite{blondel2020fast}, we have
$$
\perm = \argmax_{\perm'\in\bm{\Sigma}_d}\vp^T(\perm'\vo)
$$
\end{proof}

\subsection{Proof of \cref{prop: equivalence of evaluation}}
\label{subapp: proof of prop equivalence evaluation}
\begin{proof}
    \begin{align*}
        &\E_{p(\vp, \mW,\mTheta\vert \mD)}\left[f(\mG=\tau(\vp,\mW), \mTheta)\right]\\
        =& \int p(\vp,\mW,\mTheta, \tmW\vert \mD)f(\mG,\mTheta)d\vp d\mW d\mTheta d\tmW\\
        =&\int p(\vp, \tmW,\mTheta\vert \mD)p(\mW\vert \vp,\mTheta,\tmW,\mD)f(\mG,\mTheta)d\vp d\mW d\mTheta d\tmW\\
        =& \E_{p(\vp,\tmW,\mTheta\vert \mD)}\left[
        \frac{
        \int p(\mD\vert \vp,\mTheta,\mW)p(\vp)p(\tmW)p(\mW\vert \tmW)p(\mTheta\vert \vp,\mW)f(\mG,\mTheta)d\mW
        }{\int p(\mD\vert \vp,\mTheta,\mW)p(\vp)p(\tmW)p(\mW\vert \tmW)p(\mTheta\vert \vp,\mW)d\mW}
        \right]\\
        =&\E_{p(\vp,\tmW,\mTheta)}\left[
        \frac{
        \E_{p(\mW\vert \tmW)}\left[f(\mG,\mTheta)p(\mD,\mTheta\vert \vp,\mW)\right]
        }{
        \E_{p(\mW\vert \tmW)}\left[p(\mD,\mTheta\vert \vp, \mW)\right]
        }
        \right]
    \end{align*}
\end{proof}

\subsection{Proof of \cref{prop: joint inference gradient}}
\label{subapp: joint inference gradient}
\begin{proof}
    \begin{align*}
\nabla_\vp U(\vp,\tmW,\mTheta) &= -\nabla_{\vp} \log p(\vp,\tmW,\mTheta, \mD)\\
&=-\nabla_{\vp}\log p(\vp) - \nabla_{\vp} \log p(\tmW,\mTheta,\mD\vert \vp)\\
&=-\nabla_{\vp}\log p(\vp) - \frac{
\nabla_{\vp}\int p(\mD\vert \mW,\vp,\mTheta)p(\mTheta\vert \vp,\mW)p(\mW\vert \tmW)p(\tmW)d\mW
}{
\int p(\mD\vert \mW,\vp,\mTheta)p(\mTheta\vert \vp,\mW)p(\mW\vert \tmW)p(\tmW)d\mW
}\\
&=-\nabla_{\vp}\log p(\vp) - \frac{
\nabla_{\vp}\E_{p(\mW\vert\tmW)}\left[p(\mD\vert \mW,\vp,\mTheta)\right]
}{\E_{p(\mW\vert\tmW)}\left[p(\mD\vert \mW,\vp,\mTheta)\right]}\\
&=-\nabla_{\vp}\log p(\vp) - \nabla_{\vp} \log \E_{p(\mW\vert\tmW)}\left[p(\mD\vert \mW,\vp,\mTheta)\right]
    \end{align*}

Other gradient $\nabla_{\tmW} U$ and $\nabla_{\mTheta} U$ can be derived using the similar approach, which concludes the proof.
\end{proof}

\subsection{Proof of \cref{prop: gradient computation}}
\label{subapp: proof of proposition gradient computation}
\begin{proof}[Proof of \cref{prop: gradient computation}]
    By definition, we have easily have
    \begin{align*}
        \nabla_{\vp} U &= -\nabla_{\vp} \log p(\vp,\mW,\mTheta,\mD)\\
        &=-\nabla_{\vp}\log p(\vp,\mW) - \nabla_{\vp}\log p(\mD,\mTheta\vert \tau(\mW,\vp))\\
        &=-\nabla_{\vp}\log p(\vp,\mW) - \nabla_{\vp}\log p(\mD\vert \mTheta, \tau(\mW,\vp)) + \underbrace{\nabla_{\vp} \log p(\mTheta\vert \tau(\vp,\mW))}_{0}
    \end{align*}
Similarly, we have
\begin{align*}
    \nabla_{\mTheta} U &=  -\nabla_{\mTheta} \log p(\vp,\mW,\mTheta, \mD)\\
    &=-\nabla_{\mTheta} \log p(\mD\vert \mTheta,\tau(\mW,\vp)) - \nabla_{\mTheta}\log p(\mTheta\vert \vp,\mW) - \underbrace{\nabla_{\mTheta} \log p(\vp,\mW)}_{0}\\
    &= -\nabla_{\mTheta} \log p(\mD\vert \mTheta,\tau(\mW,\vp)) - \nabla_{\mTheta}\log p(\mTheta)
\end{align*}
\end{proof}

\subsection{Derivation of ELBO}
\label{subapp: derivation of ELBO}

\begin{align}
    \log p(\vp,\mTheta,\mD)&= \log \int p(\vp,\mTheta,\mD,\mW)d\mW \nonumber \\
    &= \log \int \frac{q_\phi(\mW\vert \vp)}{q_\phi(\mW\vert \vp)}p(\vp,\mTheta,\mD,\mW)d\mW \nonumber\\
    &\geq \int q_\phi(\mW\vert \vp)\log p(\vp,\mTheta,\mD\vert \mW)d\mW + \int q_\phi(\mW\vert \vp)\log \frac{p(\mW)}{q_\phi(\mW\vert \vp)}d\mW \label{eq: derivation of ELBO}\\
    &=\E_{q_\phi(\mW\vert \vp)}\left[\log p(\vp,\mTheta,\mD\vert \mW)\right]-\KL\left[q_\phi(\mW\vert\vp)\Vert p(\mW)\right]\nonumber
\end{align}
where the \cref{eq: derivation of ELBO} is obtained by Jensen's inequality.
\section{SG-MCMC Update}
\label{app: SG-MCMC update}

Assume we want to draw samples $\vp\sim p(\vp\vert \mD,\mW,\mTheta)\propto \exp(-U(\vp,\mW,\mTheta))$ with $U(\vp,\mW,\mTheta)=-\log p(\vp,\mW,\mTheta)$, we can compute $U$ by
\begin{equation}
    U(\vp,\mW,\mTheta) = -\sum_{n=1}^N\log p(\vx_n|\mG=\tau(\mW,\vp),\mTheta) - \log p(\vp,\mW,\mTheta)
    \label{eq: def of U}
\end{equation}
In practice, we typically use mini-batches $\mathcal{S}$ instead of the entire dataset $\mD$. Therefore, an approximation is 
\begin{equation}
    \tilde{U}(\vp,\mW,\mTheta) = -\frac{\vert \mD\vert}{\vert \mathcal{S}\vert}\sum_{n\in\mathcal{S}} \log p(\vx_n\vert \mG=\tau(\mW,\vp),\mTheta) -\log p(\vp,\mW,\mTheta)
    \label{eq: def of minibatch U}
\end{equation}
where $\vert \mathcal{S}\vert$ and $\vert \mD\vert$ are the minibatch and dataset sizes, respectively.

\cite{gong2019icebreaker} uses the preconditioning technique on \emph{stochastic gradient Hamiltonian Monte Carlo}(SG-HMC), similar to the preconditioning technique in \cite{li2016preconditioned}. In particular, they use a moving-average approximation of diagonal Fisher information to adjust the momentum.
The transition dynamics at step $t$ with EM discretization is 
\begin{align}
    &B = \frac{1}{2}l \nonumber\\
    &\mV_t = \beta_2 \mV_{t-1} + (1-\beta_2)\nabla_{\vp}\tilde{U}(\vp,\mW,\mTheta)\odot \nabla_{\vp}\tilde{U}(\vp,\mW,\mTheta)\nonumber\\
    &g_t=\frac{1}{\sqrt{1+\sqrt{\mV_t}}}\nonumber\\
&\vr_t = \beta_1\vr_{t-1} - lg_t\nabla_{\vp}\tilde{U}(\vp,\mW,\mTheta) + l\frac{\partial g_t}{\partial \vp_t} + s\sqrt{2l(\frac{1-\beta_1}{l}-B)}\eta\nonumber \nonumber \\
&\vp_t=\vp_{t-1}+lg_{t}\vr_t
\label{eq: SGMCMC update rule}
\end{align}
where $l^2$ is the learning rate; $(\beta_1,\beta_2)$ controls the preconditioning decay rate, $\eta$ is the Gaussian noise with $0$ mean and unit variance, and $s$ is the hyperparameter controlling the level of injected noise to SG-MCMC.
Throughout the paper, we use $(\beta_1, \beta_2)= (0.9,0.99)$ for all experiments.

\section{Experimental Settings}
\label{app:experiments}
\subsection{Baselines}
\label{app:baselines}
 For all the experimental settings, we compare with the following baselines: 
 \begin{itemize}
     \item Bootstrap GES (\textbf{BGES})~\cite{friedman2013data,chickering2002optimal} is a bootstrap based quasi-Bayesian approach for linear Gaussian models which first resamples with replacement data points at random and then estimates a linear SCM using the GES algorithm~\cite{chickering2002optimal} for each bootstrap set. GES is a score based approach to learn a point estimate of a linear Gaussian SCM. For all the experimental settings, we use 50 bootstrap sets. 
     \item Differentiable DAG Sampling (\textbf{DDS}) is a VI based approach to learn distribution over DAGs and a point estimate over the nonlinear functional parameters. DDS performs inference on the node permutation matrices, thus directly generating DAGs. Gumbel-sinkhorn~\cite{mena2018learning} is used for obtaining valid gradients and Hungarian algorithm is used for the straight-through gradient estimator~\cite{bengio2013estimating}. In the author provided implementation, for evaluation, a single permutation matrix is sampled and the logits of the edge beliefs are directly thresholded. In this work, in order to make the comaprison fair to Bayesian learning methods, we directly sample the binary adjacency matrix based on the edge logits.
     \item \textbf{BCD} Nets~\cite{cundy2021bcd} is a VI based fully Bayesian structure learning approach for linear causal models. BCD performs inference on both the node permutations through the Gumbel-sinkhorn~\cite{mena2018learning} operator as well as the model parameters through a VI distribution. Both DDS and BCD nets operate directly on full rank initializations to the Gumbel-sinkhorn operator, unlike our rank-1 initialization, which saves computations in practice.
     \item  \textbf{DIBS}~\cite{lorch2021dibs} uses SVGD~\cite{liu2016stein} with the DAG regularizer \cite{zheng2018dags} and bilinear embeddings to perform inference over both linear and nonlinear causal models. As our data generation process involves SCM with unequal noise variance, we extend DIBS framework with an inference over noise variance using SVGD, similar to the original paper. 
 \end{itemize}
 While DIBS and DDS can handle nonlinear parameterization, approaches like BGES and BCD, which are primarily designed for linear models still give competitive results when applied on nonlinear data. Given that there are limited number of baselines in the nonlinear case, and DIBS being the only fully Bayesian nonlinear baseline, we compare with BGES and BCD for all settings despite their model misspecification.
\subsection{Evaluation Metrics}
\label{app:metrics}
For higher dimensional settings with nonlinear models, the true posterior is intractable. While in general it is hard to evaluate the posterior inference quality in high dimensions, prior work has suggested to evaluate on proxy metrics which we adopt in this work as well~\cite{lorch2021dibs,geffner2022deep,annadani2023differentiable}. In particular, we evaluate the following metrics:
\begin{itemize}
    \item \textbf{$\E$-SHD}: Structural Hamming Distance (SHD) measures the hamming distance between graphs. In particular, it is a measure of number of edges that are to be added, removed or reversed to get the ground truth from the estimated graph. Since we have a posterior distribution $q(\mG)$ over graphs, we measure the \emph{expected} SHD: 
    \begin{equation*}
    \E\text{-SHD} \coloneqq \E_{\mG\sim q(\mG)}[\mathrm{SHD}(\mG, \mG^{{GT}})] \approx \frac{1}{N_e}\sum_{i=1}^{N_e}[\mathrm{SHD}(\mG^{(i)}, \mG^{{GT}})]~~~~,\text{with}~~~\mG^{(i)}\sim q(\mG)
\end{equation*} where $\mG^{GT}$ is the ground-truth causal graph. 
\item \textbf{Edge F1}: It is F1 score of each edge being present or absent in comparison to the true edge set, averaged over all edges.
\item \textbf{NLL}: We also measure the negative log-likelihood of the held-out data, which is also typically used as a proxy metric on evaluating the posterior inference quality \cite{gong2018meta,ma2019variational,sun2019functional}.
\end{itemize} 
The first two metrics measure the goodness of the graph posterior while the NLL measures the goodness of the joint posterior over the entire causal model.

For $d=5$ with linear models (unequal noise variance, identifiable upto MEC~\cite{peters2014identifiability, hoyer2008nonlinear}), we evaluate the following metrics:
\begin{itemize}
    \item \textbf{MMD True Posterior:} Since the true posterior is tractable, we compare the approximation with the ground truth using a Maximum Mean Discrepancy (MMD) metric~\cite{gretton2012kernel}. If $\mathrm{P}\coloneqq p(\mG\mid\mD)$ is the marginalized true posterior over graphs and $\mathrm{Q}$ is the approximated posterior over graphs, then the MMD between these two distributions is defined as:
\begin{equation*}
    \mathrm{MMD}^2(\mathrm{P},\mathrm{Q}) = \E_{\mG\sim\mathrm{P}}[k(\mG,\mG)] + \E_{\mG\sim\mathrm{Q}}[k(\mG,\mG)] - 2\E_{\mG\sim\mathrm{P},\mG'\sim\mathrm{Q}}[k(\mG,\mG')]
\end{equation*}
where $k(\mG,\mG')=1-\frac{H(\mG,\mG')}{d^2}$ is the Hamming kernel, and $H$ is the Hamming distance between $\mG$ and $\mG'$. This requires just the samples from the true posterior and the model. 
For calculating the true posterior which involves marginalization of the model parameters, appropriate prior over these parameters are required. This is ensured by using BGe score~\cite{geiger2002parameter,kuipers2014addendum} which places a Gaussian Wishart prior on the parameters. This leads to closed form marginal likelihood which is distribution equivalent, i.e. all graphs within the MEC will have the same likelihood. In addition, due to the low dimensonality ($d=5$), we can enumerate all possible DAGs and compute the normalizing constant $p(\mD)$. We refer to \cite{geiger2002parameter} for details. This metric has been used in prior work~\cite{annadani2021variational}.
\item \textbf{$\E$-CPDAG SHD:} An MEC can be represented by a Completed Partially Directed Acyclic Graph (CPDAG)~\cite{peters2017elements} which contains both directed edges and arcs (undirected edges). When causal relations between certain set of variables can be established, a directed edge is present. If there is an association between a certain set of variables for which causal direction is not identifiable, an undirected edge is present. For any graph, it has a corresponding CPDAG associated to the MEC which it belongs to. Since the ground truth graph is identifiable only upto MEC, we compare the (structural) Hamming distance between the graph posterior and the CPDAG of the ground truth. This is done by computing the $\E$-CPDAG SHD:
\begin{equation*}
    \E\text{-CPDAG SHD} \coloneqq \E_{\mG\sim q(\mG)}[\mathrm{SHD}(\mG_{\text{CPDAG}}, \mG_{\text{CPDAG}}^{{GT}})] \approx \frac{1}{N_e}\sum_{i=1}^{N_e}[\mathrm{SHD}(\mG_{\text{CPDAG}}^{(i)}, \mG_{\text{CPDAG}}^{{GT}})]
\end{equation*}
with $\mG^{(i)}\sim q(\mG)$ and $\mG_{CPDAG}^{GT}$ is the ground-truth CPDAG.
\end{itemize}

\subsection{Synthetic Data}
\label{app:synthetic_data}
As knowledge of ground truth graph is not possible in many real world settings, it is standard across causal discovery to benchmark in synthetic data settings. Following prior work, we generate synthetic data by first sampling a DAG at random from either Erdos-Rènyi (ER)~\cite{erdHos1960evolution} or Scale-Free (SF)~\cite{barabasi1999emergence} family. For $d=5$, we ensure that the graphs have $d$ edges in expectation and $2d$ edges for $d>5$. The ground truth parameters for linear functions are drawn at random from a fixed range of $[0.5, 1.5]$. For nonlinear models, the nonlinear functions are defined by randomly initialized Multi-Layer Perceptrons (MLP) with a single hidden layer of 5 nodes and ReLU nonlinearity. The variance of the exogenous Gaussian noise variable is drawn from an Inverse Gamma prior with concentration $\alpha=1.5$ and rate $\beta=1$. For $d=5$ linear case, we sample at random $N=500$ samples from the SCM for training and $N=100$ for held-out evaluation. For higher dimensional settings, we consider $N=5000$ random samples for training and $N=1000$ samples for held-out evaluation. For all settings, we evaluate on 30 random datasets.

\subsection{Hyperparameter Selection}
\label{appsubsec: hyperparameter selection}
In this section, we will give the details our how to select the hyperparameters for our method and all the baseline models. 

 We employ a cross-validation-like procedure for hyperparameter tuning in BayesDAG and DIBS to optimize MMD true posterior (for $d=5$ linear setting) and $\mathbb{E}-$SHD value (for nonlinear setting). For each ER and SF dataset with varying dimensions, we initially generate five tuning datasets. After determining the optimal hyperparameters, we fix them and evaluate the models on 30 test datasets. For DDS, we adopt the hyperparameters provided in the original paper \cite{charpentier2022differentiable}. BCD and BGES do not necessitate hyperparameter tuning since BCD already incorporates the correct prior graph for ER and SF datasets. For semi-synthetic Syntren and real world Sachs protein cells datasets, we assume the number of edges in the ground truth graphs are known and we tune our hyperparameters to produce roughly correct number of edges. BCD and DIBS also assume access to the ground truth edge number and use the graph prior to enforce the number of edges. 

\begin{table}[]
\centering
\begin{tabular}{lllll}
\hline
\multicolumn{5}{c}{\ModelName{}}                                                                                                                     \\ \hline
                                          & \multicolumn{1}{l|}{$\lambda_s$} & \multicolumn{1}{l|}{Scale $\vp$} & \multicolumn{1}{l|}{Scale $\mTheta$} & Sparse Init. \\ \hline
\multicolumn{1}{l|}{linear ER $d=5$}      & \multicolumn{1}{l|}{50}            & \multicolumn{1}{l|}{0.001}            & \multicolumn{1}{l|}{0.001}                &  False            \\
\multicolumn{1}{l|}{linear SF $d=5$}      & \multicolumn{1}{l|}{50}            & \multicolumn{1}{l|}{0.01}            & \multicolumn{1}{l|}{0.001}                &   False           \\
\multicolumn{1}{l|}{nonlinear ER $d=20$}  & \multicolumn{1}{l|}{300}         & \multicolumn{1}{l|}{0.01}        & \multicolumn{1}{l|}{0.01}            & False        \\
\multicolumn{1}{l|}{nonlinear SF $d=20$}  & \multicolumn{1}{l|}{200}         & \multicolumn{1}{l|}{0.1}         & \multicolumn{1}{l|}{0.1}             & False        \\
\multicolumn{1}{l|}{nonlinear ER $d=30$}  & \multicolumn{1}{l|}{500}         & \multicolumn{1}{l|}{1}           & \multicolumn{1}{l|}{0.01}            & False        \\
\multicolumn{1}{l|}{nonlinear SF $d=30$}  & \multicolumn{1}{l|}{300}         & \multicolumn{1}{l|}{0.01}        & \multicolumn{1}{l|}{0.01}            & False        \\
\multicolumn{1}{l|}{nonlinear ER $d=50$}  & \multicolumn{1}{l|}{500}         & \multicolumn{1}{l|}{0.01}        & \multicolumn{1}{l|}{0.01}            & True         \\
\multicolumn{1}{l|}{nonlinear SF $d=50$}  & \multicolumn{1}{l|}{300}         & \multicolumn{1}{l|}{0.1}         & \multicolumn{1}{l|}{0.01}            & False        \\
\multicolumn{1}{l|}{nonlinear ER $d=70$}  & \multicolumn{1}{l|}{700}         & \multicolumn{1}{l|}{0.1}         & \multicolumn{1}{l|}{0.01}            & True         \\
\multicolumn{1}{l|}{nonlinear SF $d=70$}  & \multicolumn{1}{l|}{300}         & \multicolumn{1}{l|}{0.01}        & \multicolumn{1}{l|}{0.01}            & False        \\
\multicolumn{1}{l|}{nonlinear ER $d=100$} & \multicolumn{1}{l|}{700}            & \multicolumn{1}{l|}{0.1}            & \multicolumn{1}{l|}{0.01}                &  False            \\
\multicolumn{1}{l|}{nonlinear SF $d=100$} & \multicolumn{1}{l|}{700}            &\multicolumn{1}{l|}{0.1 }            & \multicolumn{1}{l|}{0.01}                &  False            \\
\multicolumn{1}{l|}{SynTren}              & \multicolumn{1}{l|}{300}         & \multicolumn{1}{l|}{0.1}         & \multicolumn{1}{l|}{0.01}            & False        \\
\multicolumn{1}{l|}{Sachs Protein Cells}              & \multicolumn{1}{l|}{1200}        & \multicolumn{1}{l|}{0.1}         & \multicolumn{1}{l|}{0.01}            & False        \\ \hline
\end{tabular}
\caption{The hyperparameter selection for \ModelName{} for each setting.}
\label{tab: BayesDAG hyperparameter}
\end{table}

\begin{table}[]
\centering
\begin{tabular}{lllll}
\hline
\multicolumn{5}{c}{DIBS}                                                                                                                                  \\ \hline
                                         & \multicolumn{1}{l|}{$\alpha$} & \multicolumn{1}{l|}{$h$ latent} & \multicolumn{1}{l|}{$h_\theta$} & $h_\sigma$ \\ \hline
\multicolumn{1}{l|}{linear ER $d=5$}     & \multicolumn{1}{l|}{0.02}     & \multicolumn{1}{l|}{5}          & \multicolumn{1}{l|}{1000}       & 1          \\
\multicolumn{1}{l|}{linear SF $d=5$}     & \multicolumn{1}{l|}{0.02}     & \multicolumn{1}{l|}{15}         & \multicolumn{1}{l|}{500}        & 1          \\
\multicolumn{1}{l|}{nonlinear ER $d=20$} & \multicolumn{1}{l|}{0.02}     & \multicolumn{1}{l|}{5}          & \multicolumn{1}{l|}{1500}       & 10         \\
\multicolumn{1}{l|}{nonlinear SF $d=20$} & \multicolumn{1}{l|}{0.2}      & \multicolumn{1}{l|}{5}          & \multicolumn{1}{l|}{1500}       & 10         \\
\multicolumn{1}{l|}{nonlinear ER $d=30$} & \multicolumn{1}{l|}{0.2}      & \multicolumn{1}{l|}{5}          & \multicolumn{1}{l|}{500}        & 1          \\
\multicolumn{1}{l|}{nonlinear SF $d=30$} & \multicolumn{1}{l|}{0.2}      & \multicolumn{1}{l|}{5}          & \multicolumn{1}{l|}{1000}       & 1          \\
\multicolumn{1}{l|}{nonlinear ER $d=50$} & \multicolumn{1}{l|}{0.2}      & \multicolumn{1}{l|}{5}          & \multicolumn{1}{l|}{500}        & 10         \\
\multicolumn{1}{l|}{nonlinear SF $d=50$} & \multicolumn{1}{l|}{0.2}      & \multicolumn{1}{l|}{5}          & \multicolumn{1}{l|}{1500}       & 1          \\
\multicolumn{1}{l|}{SynTren}             & \multicolumn{1}{l|}{0.2}         & \multicolumn{1}{l|}{5}           & \multicolumn{1}{l|}{500}           &  10          \\
\multicolumn{1}{l|}{Sachs Protein Cells}             & \multicolumn{1}{l|}{0.2}         & \multicolumn{1}{l|}{5}           & \multicolumn{1}{l|}{500}           &  10          \\ \hline
\end{tabular}
\caption{The hyperparameter selection for DIBS for each setting.}
\label{tab: Dibs hyperparameter}
\end{table}

\paragraph{Network structure}
We use one hidden layer MLP with hidden size of $\max(4*d, 64)$ for the nonlinear functional relations, where $d$ is the dimensionalilty of dataset. We use \textbf{LeakyReLU} as the activation function. We also enable the \textbf{LayerNorm} and \textbf{residual connections} in the network. In particular, for variational network $\mu_\phi$ in \ModelName{}, we apply the \textbf{LayerNorm} on $\vp$ before inputting it to the network. We use 2 hidden layer MLP with size 48, \textbf{LayerNorm} and \textbf{residual connections} for $\mu_\phi$. 

\paragraph{Sparse initialization for \ModelName{}}
For \ModelName{}, we additionally allow sparse initialization by sampling a sparse $\mW$ from the $\mu_\phi$. This can be achieved by substracting a constant $1$ from the existing logits (i.e.~the output from $\mu_\phi$). 
\paragraph{Other hyperparameters}
For \ModelName{}, we run $10$ parallel SG-MCMC chains for $\vp$ and $\mTheta$. We implement an adaptive sinkhorn iteration where the iteration automatically stops when the sum of rows and columns are closed to $1$ within the threshold $0.001$ (upto a maximum of $3000$ iterations). Typically, we found this to require only around $300$ iterations. We set the sinkhorn temperature $t$ to be $0.2$. For the reparametrization of $\mW$ matrix with Gumbel-softmax trick, we use temperature $0.2$.
During evaluation, we use $100$ SG-MCMC particles extracted from the particle buffer. We use $0.0003$ for SG-MCMC learning rate $l$ and batch size $512$. We run $700$ epochs to make sure the model is fully converged.

For DIBS, we can only use $20$ SVGD particles for evaluation due to the quadratic scaling with the number of particles. We use $0.1$ for Gumbel-softmax temperature. We run $10000$ epochs for convergence. The learning rate is selected as $0.01$. 

\cref{tab: BayesDAG hyperparameter} shows the hyperparameter selection for \ModelName{}. \cref{tab: Dibs hyperparameter} shows the hyperparameter selection for DIBS.

\section{Additional Results}
\label{app:additional results}
\subsection{Walltime Comparison}
\cref{tab:walltime} presents walltime comparison of different methods. Our method converges faster while being scalable w.r.t.~DIBS, the nonlinear Bayesian causal discovery baseline. Other methods like BGES and DDS, while faster, perform much worse in terms of uncertainty quantification. In addition BGES is limited to linear model and DDS is not a fully Bayesian method.
\begin{table}[]
\centering
\caption{Walltime results (in minutes, rounded to the nearest minute) of the runtime of different approaches on a single 40GB A100 NVIDIA GPU. The N/A fields indicate that the corresponding method cannot be run within the memory constraints of a single GPU.}
\label{tab:walltime}
\begin{tabular}{l|cccc}
\hline
             & \multicolumn{1}{c|}{d=30} & \multicolumn{1}{c|}{d=50} & \multicolumn{1}{c|}{d=70} & d=100 \\ \hline
BaDAG (\textbf{Ours})(Bayesian, Nonlinear) & 171                       & 238                       & 261                       & 448   \\
DIBS (Bayesian, Nonlinear)        & 187                       & 350                       &      N/A                     &     N/A  \\ \hline
BGES (Quasi-Bayesian, Linear)       & 2                         & 3                         & 6                         & 11    \\
BCD (Bayesian, Linear)         & 252                       & 328                       & 418                       & 600   \\
DDS  (Quasi-Bayesian, Nonlinear)        & 92                        & 130                       & 174                       &  N/A   \\
\hline
\end{tabular}
\end{table}
\subsection{Performance with higher dimensional datasets}
\label{appsubsec: higher dimensional datasets}
\begin{figure}
    \centering
    \includegraphics[width=0.75\linewidth]{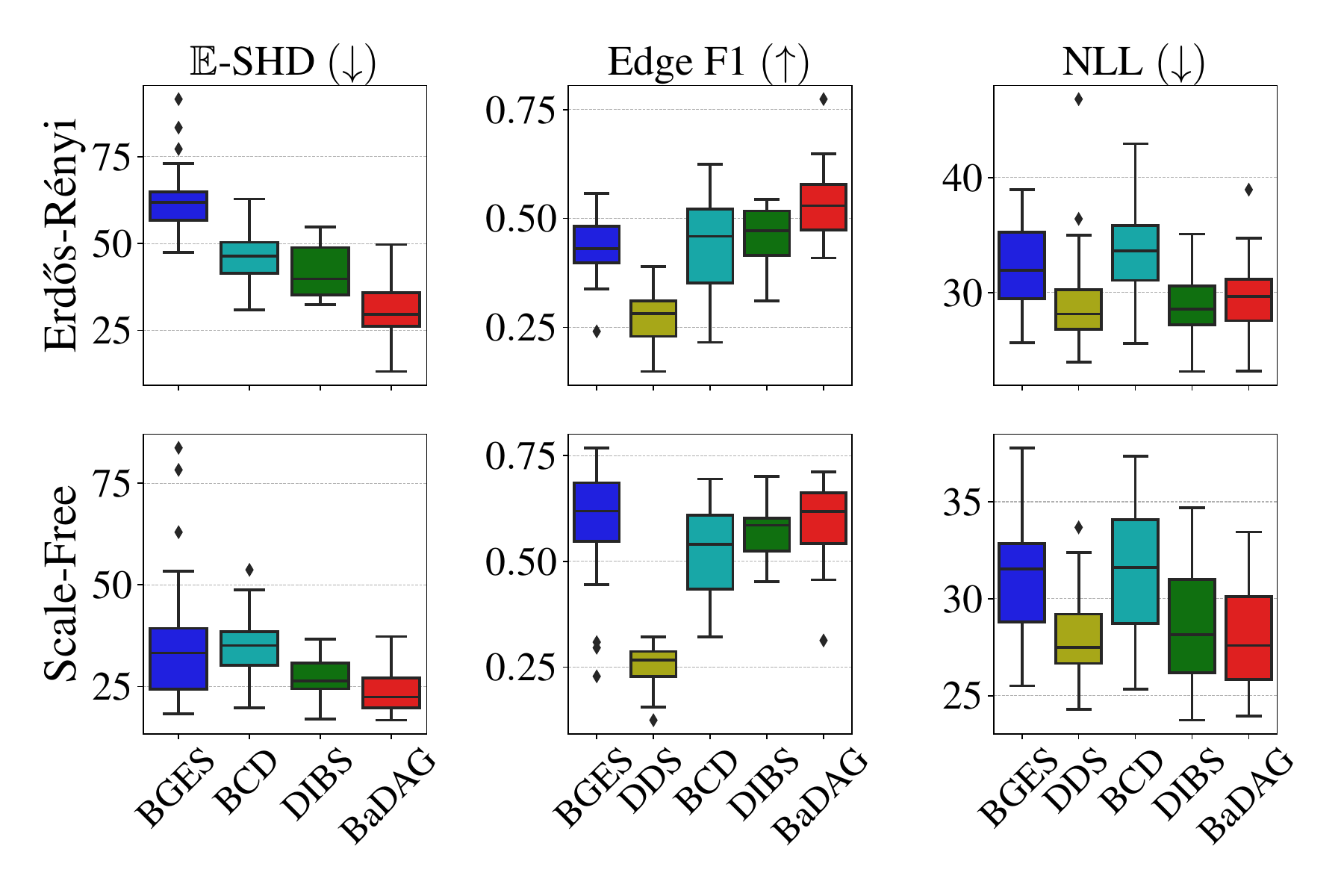}
    \caption{Posterior inference of both graph and functional parameters on synthetic datasets of nonlinear causal models with $d=20$ variables. \ModelName~gives best results across all metrics. $\downarrow$ denotes lower is better and $\uparrow$ denotes higher is better. For the sake of clarity, DDS has been omitted for $\E$-SHD due to its significantly inferior performance on this metric.}
    \label{fig:nonlinear_20_40}
\end{figure}

\begin{figure}
    \centering
    \includegraphics[width=0.75\linewidth]{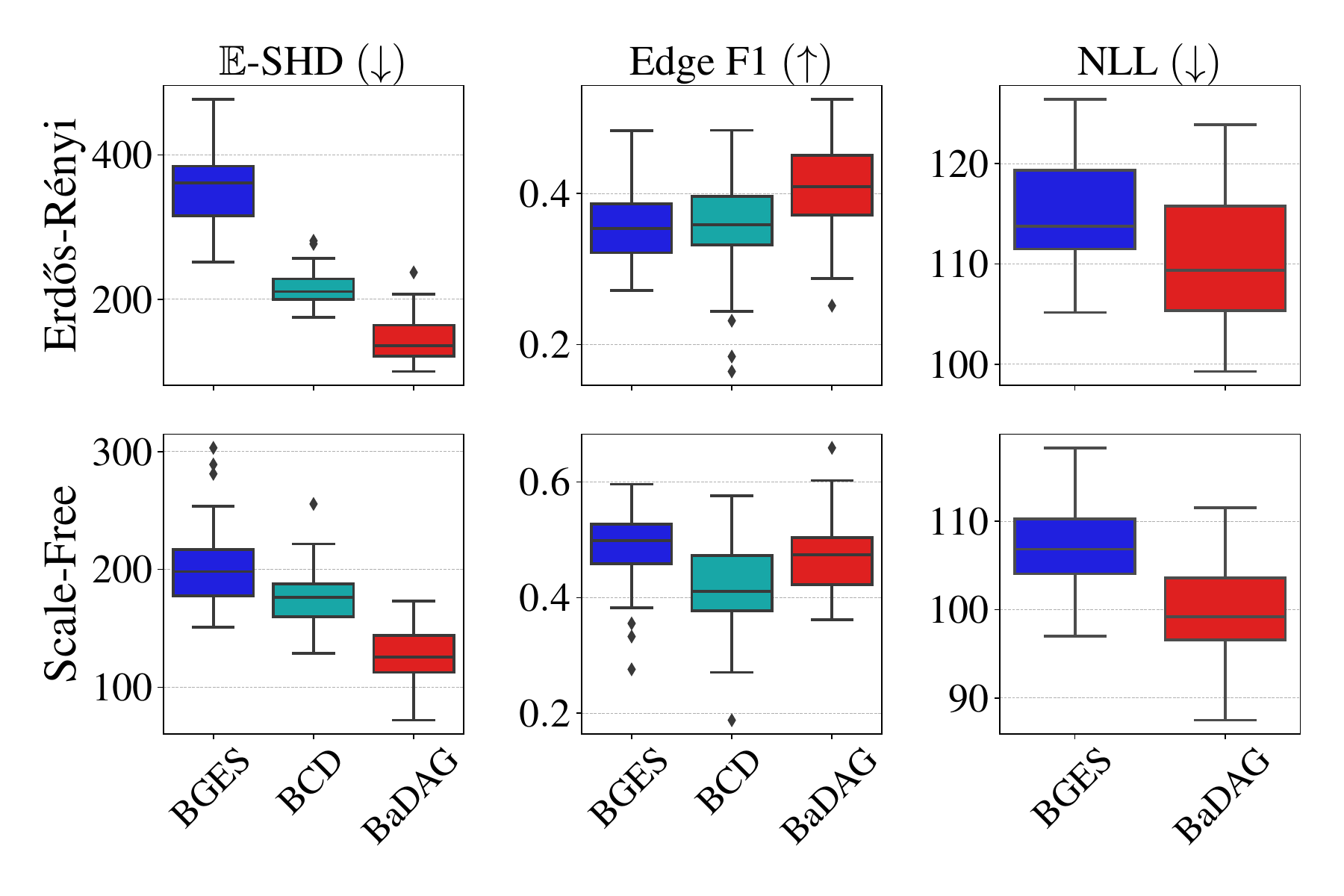}
    \caption{Posterior inference of both graph and functional parameters on synthetic datasets of nonlinear causal models with $d=70$ variables. \ModelName~gives best results across most metrics. $\downarrow$ denotes lower is better and $\uparrow$ denotes higher is better. As DIBS and DDS are computationally prohibitive to run for this setting, it has been omitted. BCD has been omitted for NLL as we observed that it performs significantly worse. }
    \label{fig:nonlinear_70_140}
\end{figure}

\begin{figure}
    \centering
    \includegraphics[width=0.75\linewidth]{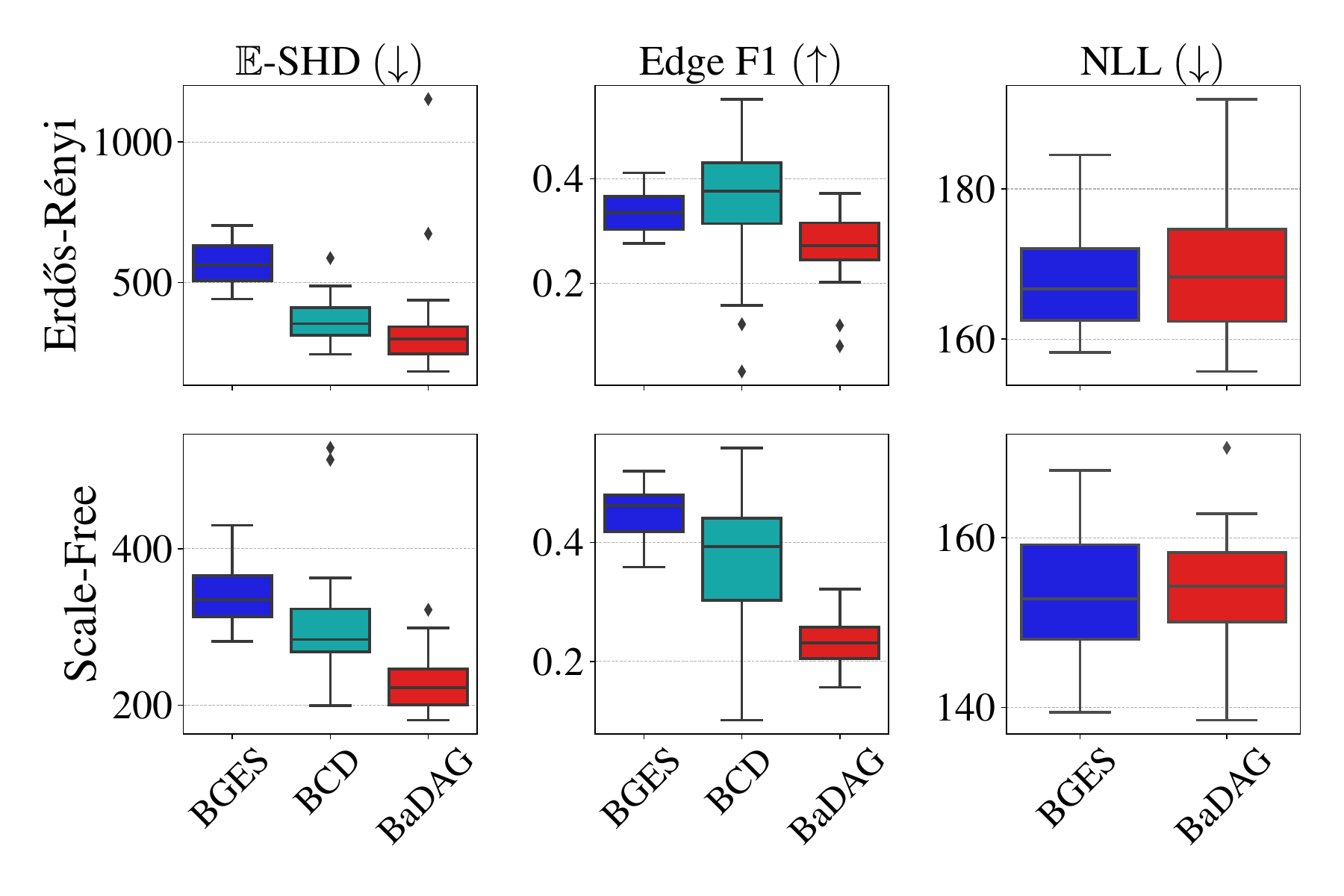}
    \caption{Posterior inference of both graph and functional parameters on synthetic datasets of nonlinear causal models with $d=100$ variables. \ModelName~gives best results across $\E$-SHD, comparable across NLL but slightly worse for Edge F1. $\downarrow$ denotes lower is better and $\uparrow$ denotes higher is better. As DIBS and DDS are computationally prohibitive to run for this setting, it has been omitted. BCD has been omitted for NLL as we observed that it performs significantly worse. }
    \label{fig:nonlinear_100_200}
\end{figure}

Full results for all the metrics for settings $d=20$, $d=70$ and $d=100$ for nonlinear settings are presented in \cref{fig:nonlinear_20_40}, \cref{fig:nonlinear_70_140} and \cref{fig:nonlinear_100_200}.
\subsection{Performance of SG-MCMC with Continuous Relaxation}
\label{appsubsec: fully SG-MCMC performance}
\begin{figure}
    \centering
    \includegraphics[width=0.85\linewidth]{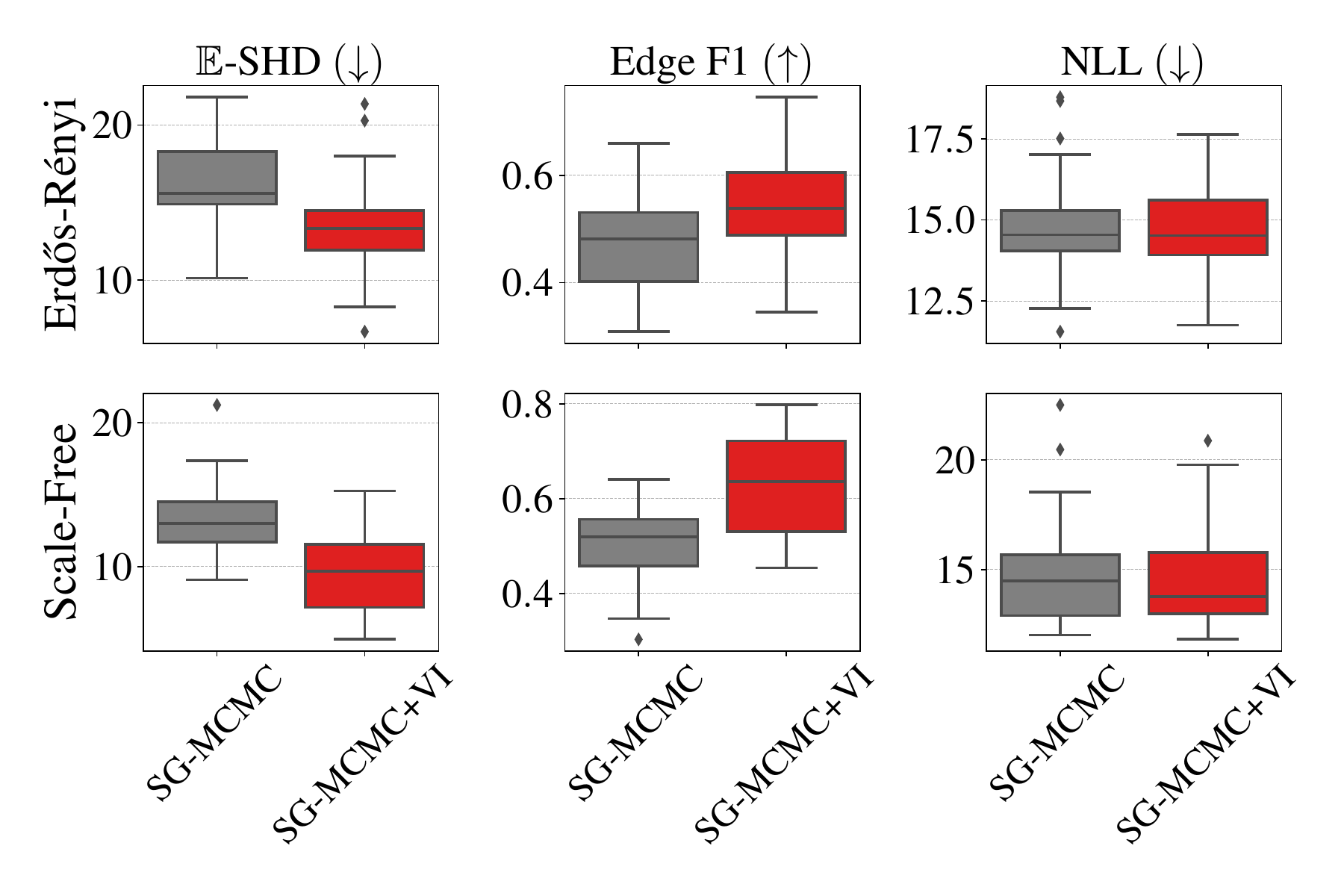}
    \caption{Performance comparison of SG-MCMC+VI v.s. fully SG-MCMC with $\tmW$ for $d=10$ variables.}
    \label{fig:sg-mcmc+vi vs sg-mcmc 10}
\end{figure}

We compare the performance of SG-MCMC+VI and SG-MCMC with $\tmW$ on $d=10$ ER and SF graph settings. \cref{fig:sg-mcmc+vi vs sg-mcmc 10} shows the performance comparison. We can observe that SG-MCMC+VI generally outperforms its counterpart in most of the metrics. We hypothesize that this is because VI network $\mu_\phi$ couples $\vp$ and $\mW$. This coupling effect is crucial since the changes in $\vp$ results in the change of permutation matrix, where the $\mW$ can immediately respond to this change through $\mu_\phi$. On the other hand, $\tmW$ can only respond to this change through running SG-MCMC steps on $\tmW$ with fixed $\vp$. In theory, this is the most flexible approach since this coupling do not requires parametric form like $\mu_\phi$. However in practice, we cannot run many SG-MCMC steps with fixed $\vp$ for convergence, which results in the inferior performance. 

\section{Code and License}
For the baselines, we use the code from the following repositories:
\begin{itemize}
    \item BGES: We use the code from \cite{agrawal2019abcd} from the repository \href{https://github.com/agrawalraj/active_learning}{https://github.com/agrawalraj/active\_learning} (No license included).
    \item DDS: We use the code from the official repository \href{https://github.com/sharpenb/Differentiable-DAG-Sampling}{https://github.com/sharpenb/Differentiable-DAG-Sampling} (No license included).
    \item BCD: We use the code from the official repository \href{https://github.com/ermongroup/BCD-Nets}{https://github.com/ermongroup/BCD-Nets} (No license included).
    \item DIBS: We use the code from the official repository \href{https://github.com/larslorch/dibs}{https://github.com/larslorch/dibs} (MIT license).
\end{itemize}
Additionally for the Syntren~\cite{van2006syntren} and Sachs Protein Cells~\cite{sachs2005causal} datasets, we use the data provided with repository \href{https://github.com/kurowasan/GraN-DAG}{https://github.com/kurowasan/GraN-DAG} (MIT license).

\section{Broader Impact Statement}

This work is concerned with understanding cause and effects from data, which has potential applications in empirical sciences, economics and machine perception. Understanding causal relationships can improve fairness in decision making, understand biases which might be present in the data and answering causal queries. As such, we envision this line of work to not have any significant negative impact.
\end{document}